\documentclass[journal]{IEEEtran}
\usepackage{cite}
\ifCLASSINFOpdf
\else
\fi
\usepackage{algorithmic}
\ifCLASSOPTIONcompsoc
 \usepackage[caption=false,font=normalsize,labelfont=sf,textfont=sf]{subfig}
\else
 \usepackage[caption=false,font=footnotesize]{subfig}
\fi
\usepackage{amsmath,amssymb,amsfonts}
\usepackage{amssymb}
\usepackage{breqn}
\usepackage{hyperref} \hypersetup{hidelinks}
\usepackage{cleveref}
\usepackage[lined,ruled,linesnumbered,commentsnumbered]{algorithm2e}
\usepackage{booktabs}
\usepackage{cases}
\usepackage{graphicx}
\usepackage{multirow}
\usepackage{mathrsfs}
\usepackage{textcomp}
\usepackage{xcolor}
\usepackage{boxedminipage}
\usepackage{mathrsfs}
\usepackage{enumerate}
\usepackage{threeparttable}

\begin{document}
\title{Efficiently Tackling Million-Dimensional Multiobjective Problems: A Direction Sampling and Fine-Tuning Approach}

\author{Haokai~Hong,
        Min~Jiang,~\IEEEmembership{Senior Member,~IEEE,}
        Qiuzhen Lin,~\IEEEmembership{Senior Member,~IEEE,}
        and~Kay Chen~Tan,~\IEEEmembership{Fellow,~IEEE}
    \thanks{© 2024 IEEE. Personal use of this material is permitted.  Permission from IEEE must be obtained for all other uses, in any current or future media, including reprinting/republishing this material for advertising or promotional purposes, creating new collective works, for resale or redistribution to servers or lists, or reuse of any copyrighted component of this work in other works.}
    \thanks{H. Hong and M. Jiang are with the Department of Artificial Intelligence, Key Laboratory of Digital Protection and Intelligent Processing of Intangible Cultural Heritage of Fujian and Taiwan, Ministry of Culture and Tourism, School of Informatics, Xiamen University, Fujian, China, 361005. {\itshape (Corresponding authors: M. Jiang)}}
    \thanks{Q. Z. Lin is with the College of Computer Science and Software Engineering, Shenzhen University, China.}
    \thanks{K. C. Tan is with the Department of Computing, Hong Kong Polytechnic University, Hong Kong SAR.}
}

\markboth{Journal of \LaTeX\ Class Files,~Vol.~14, No.~8, August~2015}%
{Shell \MakeLowercase{\textit{et al.}}: Bare Demo of IEEEtran.cls for IEEE Journals}

\maketitle

\begin{abstract}
We define very large-scale multiobjective optimization problems as optimizing multiple objectives (VLSMOPs) with more than 100,000 decision variables. These problems hold substantial significance, given the ubiquity of real-world scenarios necessitating the optimization of hundreds of thousands, if not millions, of variables. However, the larger dimension in VLSMOPs intensifies the curse of dimensionality and poses significant challenges for existing large-scale evolutionary multiobjective algorithms, rendering them more difficult to solve within the constraints of practical computing resources. To overcome this issue, we propose a novel approach called the very large-scale multiobjective optimization framework (VMOF). The method efficiently samples general yet suitable evolutionary directions in the very large-scale space and subsequently fine-tunes these directions to locate the Pareto-optimal solutions. To sample the most suitable evolutionary directions for different solutions, Thompson sampling is adopted for its effectiveness in recommending from a very large number of items within limited historical evaluations. Furthermore, a technique is designed for fine-tuning directions specific to tracking Pareto-optimal solutions. To understand the designed framework, we present our analysis of the framework and then evaluate VMOF using widely recognized benchmarks and real-world problems spanning dimensions from 100 to 1,000,000. Experimental results demonstrate that our method exhibits superior performance not only on LSMOPs but also on VLSMOPs when compared to existing algorithms.
\end{abstract}

\IEEEpeerreviewmaketitle
\section{Introduction}
Multiobjective optimization problems (MOPs) are characterized by multiple conflicting objectives and find extensive applications in various real-world domains~\cite{996017, 6848830}. Within scientific and engineering fields, a wide range of MOPs exists, featuring a substantial number of decision variables. These problems are commonly referred to as large-scale multiobjective optimization problems (LSMOPs)~\cite{tian2021evolutionary, hong2021evolutionary}. The `curse of dimensionality' is a significant challenge faced by conventional multiobjective evolutionary algorithms (MOEAs) when tackling LSMOPs. This challenge arises from the exponential growth of the decision space, leading to slower convergence compared to mathematical programming methods~\cite{tian2021evolutionary,9047876}.
\par
To mitigate the curse of dimensionality, considerable efforts have been made in recent years to tailor massive multiobjective evolutionary algorithms specifically for LSMOPs~\cite{2008Large, 8720021, 8765790, 9047876, 9870430}. One notable example is the CCGDE3 algorithm proposed by Antonio \emph{et al.}~\cite{6557903}, which introduced decision variable grouping by maintaining multiple subgroups of equal length. Zille \emph{et al.} developed the WOF~\cite{RN89}, employing grouping mechanisms and a population-based optimizer for both original and weight variables. Another innovative approach is LMOCSO~\cite{8681243} by Tian \emph{et al.}, which designed a novel search strategy that utilizes the competitive swarm optimizer.
\par
While research on LSMOPs has made substantial strides, the majority of existing methodologies are primarily tailored to tackle problems with dimensions fewer than 100,000. Nonetheless, it is not uncommon to encounter MOPs with decision variables exceeding a hundred thousand. For example, in contemporary power delivery systems, voltage transformers display time-varying characteristics, leading to a time-varying ratio error estimation problem that involves more than 100,000 decision variables~\cite{8962275}. Such high-dimensional problems surpass the capabilities of current algorithms.
\par
In light of the preceding discussion, we propose a new class of optimization problems, termed Very Large-Scale Multiobjective Optimization Problems (VLSMOPs), characterized by more than 100,000 decision variables. The primary challenge in tackling VLSMOPs arises from the tension between the limited computational resources typically available and the vast search space, which complicates the task of locating the Pareto-optimal frontier (PF) - a manifestation of the intensified curse of dimensionality~\cite{donoho2000} compared to traditional LSMOPs. Therefore, the crux of solving VLSMOPs hinges on the efficient identification of evolutionary directions and the effective exploration of the PF.
\par
\begin{figure*}[ht]
    \includegraphics[width=1\hsize]{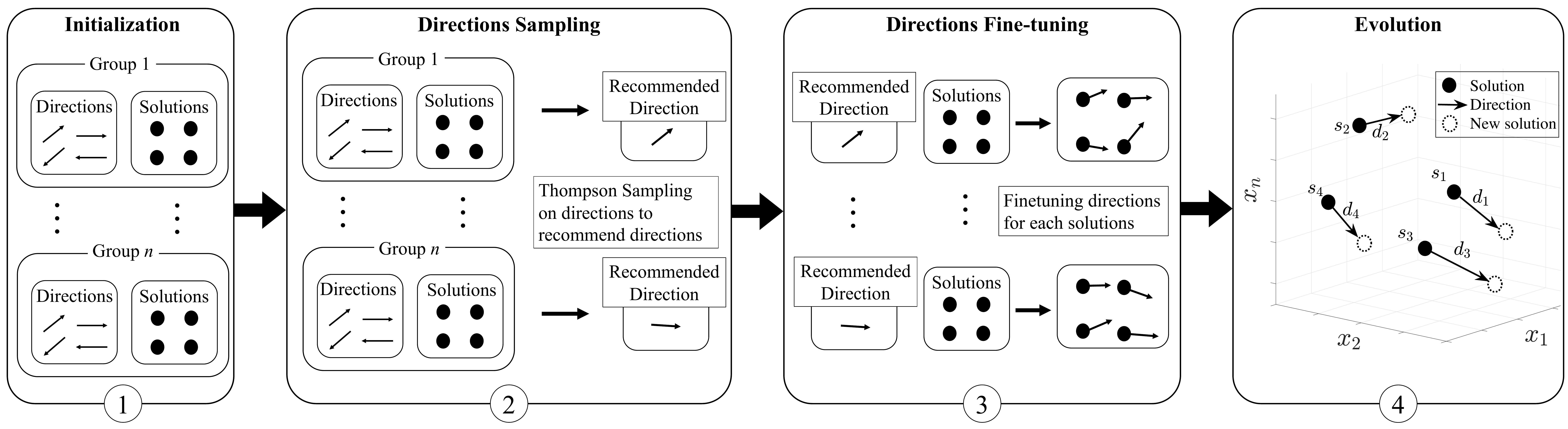}
    \caption{Illustration of the proposed very large-scale multiobjective optimization framework (VMOF). 1. Initialize a set of directions and solutions and randomly partition them into several groups. 2. Sample and recommend a direction for each group. 3. Fine-tune directions for all solutions based on recommended directions. 4. Evolve each solution according to fine-tuned directions.}
    \label{fig: VMOF}
\end{figure*}
\par
To tackle the challenges posed by VLSMOPs, we propose a specialized framework called the Very Large-Scale Multiobjective Optimization Framework (VMOF), which leverages the techniques of evolution directions sampling and fine-tuning. Our framework is specifically designed to address the curse of dimensionality encountered in VLSMOPs.
\par
Firstly, the evolution directions sampling technique draws inspiration from Thompson sampling, a method commonly employed in recommender systems (RS)~\cite{BOBADILLA2013109} to solve the cold-start problem~\cite{10.1145/3446427}. Thompson sampling is especially efficient in recommending the optimal choice from a vast pool of items based on limited historical interactions, which is consistent with the need to solve VLSMOPs. Therefore, we develop a method that establishes a distribution of evolution directions in the very large-scale search space and subsequently samples from this distribution to recommend the most favorable evolution direction.
\par
However, having suitable yet general evolutionary directions alone is insufficient for locating the Pareto-optimal set (PS) within the vast decision space of VLSMOPs. To ensure the accurate and efficient search of Pareto-optimal solutions, we introduce a directions fine-tuning algorithm that calculates refined directions for each solution based on the sampled directions to better approximate the PS. Fig.~\ref{fig: VMOF} provides an illustration of the proposed framework.
\begin{enumerate}
    \item In this work, we introduce an innovative framework specifically designed to tackle the complexities inherent in very large-scale multiobjective optimization problems (VLSMOPs), a common occurrence in practical applications. Our comprehensive experimental results confirm the ability of the proposed algorithm to solve VLSMOPs on a scale of up to one million.
    \item To tackle VLSMOPs, we propose a new method that employs Thompson sampling to sample and recommend directions. This method enables our proposed framework, VMOF, to suggest general yet suitable evolutionary directions within the extensive search space typical of VLSMOPs, even when facing limitations in function evaluations.
    \item Additionally, to facilitate the exploration of the expansive decision space intrinsic to VLSMOPs, we propose a direction fine-tuning method. This approach utilizes the sampled directions to steer each solution toward the approximation of Pareto-optimal solutions.
\end{enumerate}
\par
The remainder of this paper is organized as follows. Tn Section \ref{sec: Preliminaries-and-Related}, we define VLSMOPs and present related works. The details of the proposed approach are presented in Section \ref{sec: VMOF}. We present some analyses on the designed framework in Section \ref{sec: ana}. In Section \ref{sec:exp}, we conduct a series of experimental comparisons between our proposed method and several state-of-the-art large-scale MOEAs. Finally, conclusions are drawn and future work is discussed in Section~\ref{sec:cl-fw}.
\section{Preliminary Studies and Related Work}
\label{sec: Preliminaries-and-Related}
\subsection{Very large-scale Multiobjective Optimization}
The mathematical form of LSMOPs is as follows:
\begin{equation}
\begin{aligned}
\textbf{min}\ \boldsymbol F(\boldsymbol x) =& <f_1(\boldsymbol x), f_2(\boldsymbol x),\dots,f_m(\boldsymbol x)>\\
& \textrm{s.t.}~ \boldsymbol x \in \Omega
\end{aligned}
\end{equation}
where $\boldsymbol x=(x_1,x_2,...,x_n)$ is $n$-dimensional decision vector, $\boldsymbol F=(f_1,f_2,\dots,f_m)$ is $m$-dimensional objective vector.
\par
Suppose $\boldsymbol x_1$ and $\boldsymbol x_2$ are two solutions of an MOP, solution $\boldsymbol x_1$ is known to Pareto dominate solution $\boldsymbol x_2$ (denoted as $\boldsymbol x_1 \prec \boldsymbol x_2$), if and only if $f_i(\boldsymbol x_1) \leqslant f_i(\boldsymbol x_2) (\forall i = 1,\dots,m)$ and there exists at least one objective $f_j (j \in \{1, 2, \dots , m\})$ satisfying $f_j(\boldsymbol x_1) < f_j(\boldsymbol x_2)$. The collection of all the Pareto optimal solutions in the decision space is called the Pareto optimal set (PS), and the projection of PS in the objective space is called the Pareto optimal front (PF).
\par
It should be noted that conventional large-scale optimization problems typically regard the number of decision variables $n$ as exceeding $100$ \cite{8681243}. Nonetheless, the decision variables in numerous real-world problems, such as neural architecture search, critical node detection, and the vehicle routing problem, can easily surpass $100,000$. Consequently, we introduce the concept of \textbf{V}ery \textbf{L}arge-\textbf{S}cale \textbf{M}ultiobjective \textbf{O}ptimization \textbf{P}roblems (VLSMOPs), where $n$ exceeds $100,000$.
\subsection{Related Works}
Large-scale MOEAs are categorized into three classes: decision variable grouping-based, decision space reduction-based, and novel search strategy-based approaches \cite{tian2021evolutionary}.
\subsubsection{Decision Variable Grouping Based}
The first category is based on decision variable grouping. CCGDE3, proposed by Antonio \emph{et al.} \cite{6557903}, was the first multiobjective evolutionary algorithm for solving LSMOPs. It maintains numerous distinct subgroups and optimizes LSMOP in a divide-and-conquer manner.
Afterward, random grouping \cite{10.1145/3319619.3322068} and differential grouping \cite{10.1145/3205455.3205491} techniques were introduced to solve LSMOPs.
\par
Another subcategory in decision variable grouping-based methods is the variable analysis based \cite{8416294}. MOEA/DVA was presented by Ma \emph{et al.} in \cite{7155533} to partition the decision variables by analyzing their control property. To partition the decision variables more generically, Zhang \emph{et al.} \cite{RN82} proposed a large-scale evolutionary algorithm based on a decision variable clustering approach.
\par
Grouping decision variables is an effective way to solve LSMOP, but the connections between large-scale decision variables may be incorrectly identified~\cite{RN82}, which may lead to local optima. For very large-scale decision variables, such a shortcoming may further worsen the performance.
\subsubsection{Decision Space Reduction Based}
The second category is based on dimensionality reduction \cite{10.5555/3298239.3298367}. Principal component analysis was suggested to solve LSMOPs by Liu \emph{et al.} in \cite{LIU2020106120}, which is PCA-MOEA. Furthermore, unsupervised neural networks were used by Tian \emph{et al.} to solve large-scale problems by learning the Pareto-optimal subspace \cite{9047876}.
\par
In this category, problem transformation-based algorithms have been suggested to solve LSMOPs. Zille \emph{et al.} \cite{RN89} proposed a weighted optimization framework (WOF) that was intended to serve as a generic method that can be used with any population-based metaheuristic algorithm. Besides, He \emph{et al.} \cite{RN88} proposed a large-scale multiobjective optimization framework as a general framework, which reformulated the original LSMOP as a low-dimensional single-objective optimization problem.
\par
Methods based on decision space reduction may lose important information about the original search space after reduction. When solving VLSMOPs, dimensionality reduction will lose more search space, therefore the original search space cannot be fully explored~\cite{9723458} leading to poor performance.
\subsubsection{Novel Search Strategy Based}
The third category of research is centered around novel search strategies \cite{YI2020470, 9138459}. Tian \emph{et al.} introduced a competitive swarm optimizer for efficient search solutions to LSMOPs. Concurrently, Hong \emph{et al.} developed a scalable algorithm with an improved diversification mechanism \cite{8533425}. More recently, Wang \emph{et al.} \cite{9552479} presented a generative adversarial network-based manifold interpolation framework for learning the manifold and generating high-quality solutions. Hong \emph{et al.} put forth a model that leverages a trend prediction model and a generating-filtering strategy to address LSMOPs. In a different approach, Qin \emph{et al.} \cite{9367299} proposed a large-scale multiobjective evolutionary algorithm using direct sampling. Contrary to this, we employ Thompson sampling to select the optimal direction, demonstrating both theoretically and empirically that our method is more effective for very large-scale decision variables.
\par
Numerous algorithms have been proposed over the years to solve LSMOPs. And larger-scale MOPs have drawn some attention. For example, Tian \emph{et al.} \cite{9754325} proposed a method for solving super-large-scale sparse multiobjective optimization problems. However, the designed algorithm requires that the decision variables of the problem are sparse, that is, there are a large number of 0 in the decision variables of the problem, which limits the algorithm. The work of Li \emph{et al.} \cite{LI2022101181}, explores higher dimensional MOPs, with a primary emphasis on improving the computational speed for solving LSMOPs. However, there is a dearth of comprehensive analyses and experiments on their proposed method. Notably, VLSMOPs have garnered interest in this research field. However, the topic remains relatively understudied.
\subsection{Thompson sampling}
Thompson sampling (TS) \cite{NIPS2011_e53a0a29} is a commonly used strategy in recommender systems. Thompson sampling was first proposed in 1933 \cite{10.2307/2332286, 10.2307/2371219} for allocating experimental effort in two-armed bandit problems arising in clinical trials. Thompson sampling strikes a balance between statistical efficiency and algorithmic complexity \cite{10.1561/2200000070}, and is widely adopted in Internet advertising \cite{10.1145/2556195.2556252}, recommender systems \cite{10.5555/2969239.2969384} and hyperparameter tuning \cite{pmlr-v84-kandasamy18a}. In evolutionary computation, Sun \emph{et al.} \cite{10.1007/978-3-030-58115-2_19} proposed a new adaptive operator selection mechanism for multiobjective evolutionary algorithms based on dynamic Thompson sampling.
\par
To address VMOPs, the key lies in striking a balance between exploration and exploitation within a very large-scale search space. The upper-confidence bound (UCB), a classical method for achieving this balance, accommodates both statistical and computational efficiency. However, this leads to UCB algorithms sacrificing statistical efficiency~\cite{10.1561/2200000070}. In contrast, Thompson sampling has been proven effective in solving bandit problems with high-dimensional features~\cite{pmlr-v202-chakraborty23b}. Therefore, our framework adopts Thompson sampling, which enables the algorithm to avoid direct search in the high-dimensional decision space, instead obtaining directions that can efficiently guide the evolution of the population.
\section{Sampling and Fine-tuning based Framework for Very Large-scale Multiobjective Optimization}
\label{sec: VMOF}
\subsection{Overview}
The presented framework contains two main components: directions sampling and directions fine-tuning as depicted in Fig.~\ref{fig: VMOF}. These two components aim to sample general yet suitable directions and fine-tune them for cooperatively solving VLSMOPs.
\par
Firstly, our proposal employs Thompson sampling to sample and recommend evolutionary directions within an extensive search space. Thompson sampling maintains a probabilistic distribution for each evolution direction in the large-scale space and selects samples from these distributions to make recommendations within limited function evaluations.
\par
Secondly, to mitigate the issue of accurately approximating Pareto-optimal solutions in very large-scale decision space, it is crucial to search as finely as possible. To achieve this, we introduce an algorithm that fine-tunes an evolution direction for each solution. This algorithm generates a set of directions based on the recommended directions for each solution.
\subsection{Evolution Directions Sampling}
\begin{figure*}[ht]
    \centering
    \subfloat[]{\includegraphics[width=0.2\hsize]{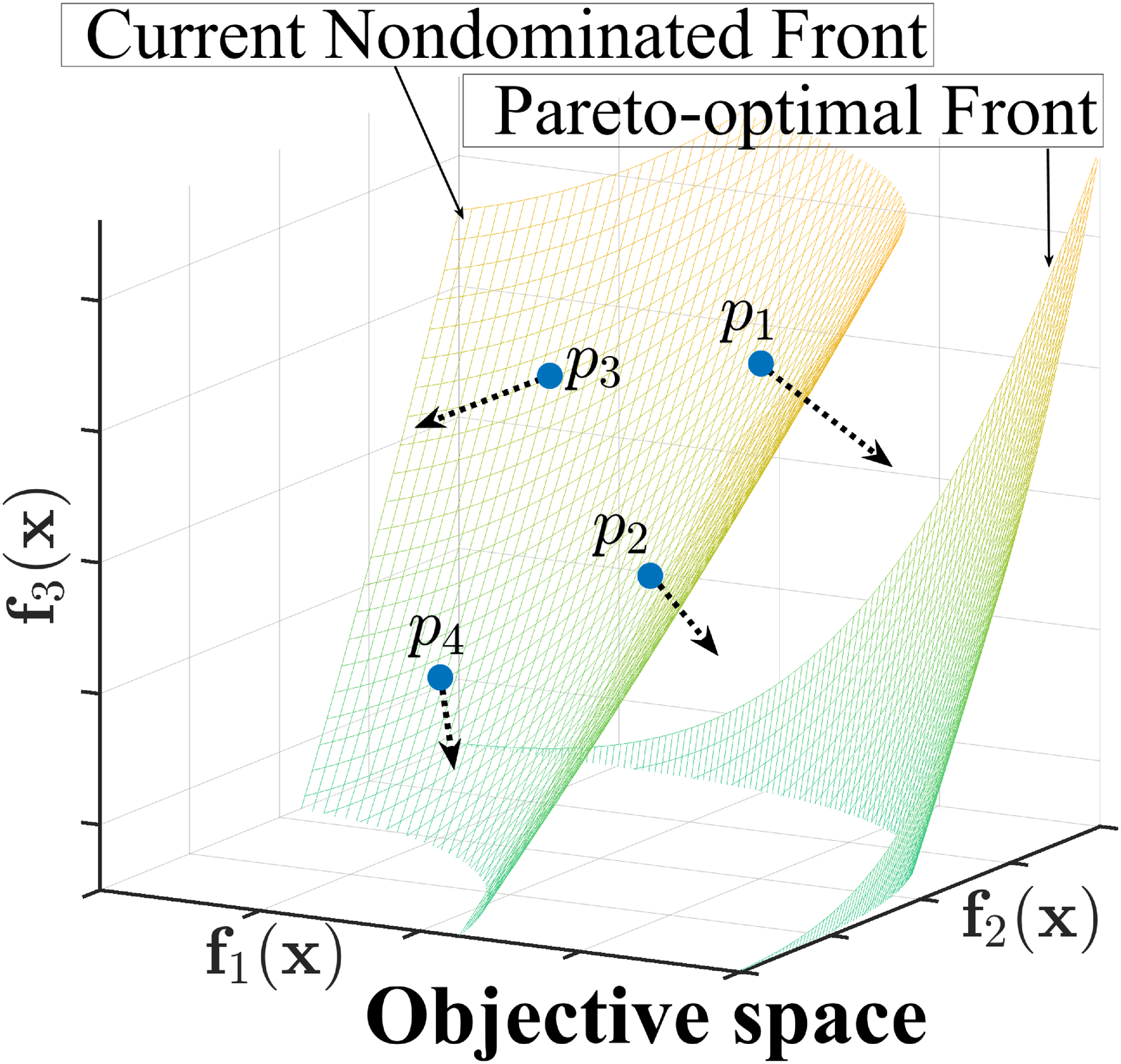}}
    \subfloat[]{\includegraphics[width=0.78\hsize]{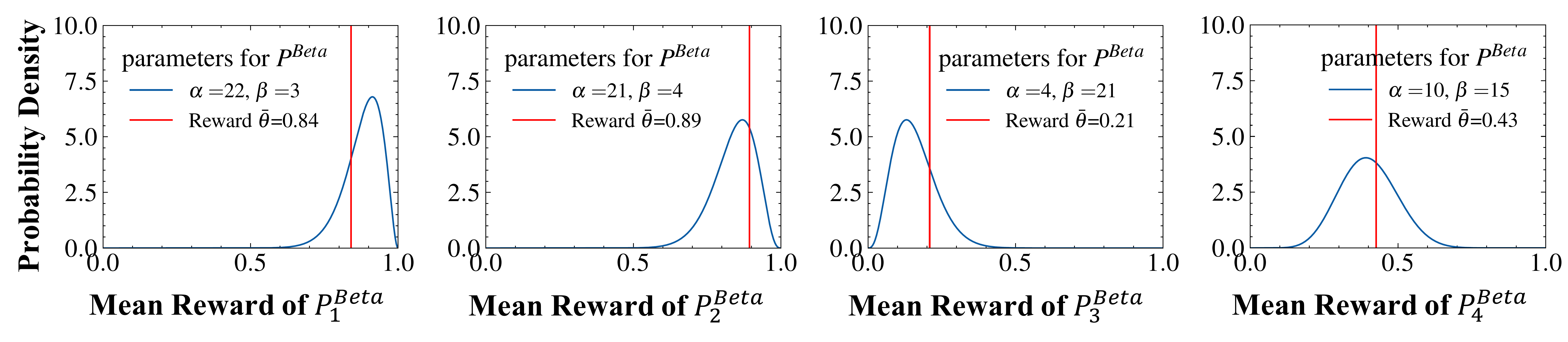}}
    \caption{This illustration depicts the process of evolution direction sampling based on Thompson sampling. Initially, each individual possesses a unique direction ($\boldsymbol d_i$) within the decision space, guiding the individual through a single evolution. Subsequently, each individual ($\boldsymbol p_i$) undergoes evolution and evaluation within the objective space. Based on the evaluation results, the algorithm updates the corresponding $\alpha$ and $\beta$ parameters for each direction. Notably, $\alpha$ and $\beta$ represent the number of times an individual has evolved and degenerated along direction $\boldsymbol d_i$, respectively. The algorithm then calculates the $P^{Beta}(\theta)$ distribution for each direction (represented by blue lines in $P^{Beta}_i$), and samples the mean reward $\bar \theta_i$ from the $P^{Beta}(\theta)$ distribution (indicated by the red lines in $P^{Beta}_i$). The direction corresponding to the distribution with the highest mean reward $\bar \theta_i$ is selected. In this example, the sampled direction is $\boldsymbol d_1$, as determined by $P^{Beta}_1$.}
    \label{fig: TSonDir}
\end{figure*}
For solutions $\mathcal S = \{ \boldsymbol s_i \}$ with dimension $d$, it is impossible to locate directions $\mathcal D = \{\boldsymbol d_j\}$ toward the PS in the whole search space since it is very large scale. Therefore, we transform the direction optimization into a cold-start problem and apply Thompson sampling to recommend the best directions.
\par
The evolution direction for a solution is defined as $\boldsymbol d~(\boldsymbol d = (d_1, d_2, \dots, d_d))$. The algorithm will recommend directions for solutions and then fine-tune directions for each solution.
\par
The cold-start problem considers allocating a fixed set of limited resources among large-scale items $\mathcal I = \{i_1,\dots, i_k\}$ that maximizes the expected return for users $\mathcal U = \{u_1, \dots,  u_l\}$. In the proposed framework, each solution $\boldsymbol s_i$ is regarded as a user $u_i$, and an item $i_j$ is the direction $\boldsymbol d_j$ to guide the evolution of the solution. The response of user $u_i$ to the item $i_j$, or denoted as the reward $r_{ij}$ is the quality of solution $\boldsymbol s_i$ after evolving along the direction $\boldsymbol d_j$.
\par
Thompson sampling is an efficient algorithm for solving the cold-start problem. In Thompson sampling, each item is assigned an independent prior reward distribution of $\theta_j, j \in \{1,\dots, k\}$. In particular, a Beta distribution with parameters $\boldsymbol \alpha = (\alpha_1,\dots, \alpha_k)$ and $\boldsymbol \beta = (\beta_1,\dots,\beta_k)$ is used to represent this prior distribution:
\begin{equation}
    \begin{aligned}
    \label{equ: beta}
    P^{Beta}(\theta_j) = \frac {\Gamma(\alpha_j+\beta_j)} {\Gamma(\alpha_j)\Gamma(\beta_j)} \theta_j^{\alpha_j-1}(1-\theta_j)^{\beta_j-1}
    \end{aligned}
\end{equation}
where $\alpha_j$ and $\beta_j$ are the parameters associated with the $j$-th item $i_j$, and $\Gamma(\cdot)$ is the gamma function. For a given item $i_j$ for user $u_i$, if its reward for user $u_i$ is $r_{ij} = 1$, then $\alpha_j$ and $\beta_j$ are updated as:
\begin{equation}
    \begin{aligned}
    \label{equ:update}
    \alpha_j = \alpha_j + 1, \beta_j = \beta_j
    \end{aligned}
    \end{equation}
    otherwise, if $r_{ij} = 0$ then we have:
    \begin{equation}
    \begin{aligned}
    \label{equ:update2}
    \alpha_j = \alpha_j, \beta_j = \beta_j + 1.
    \end{aligned}
\end{equation}
\par
\begin{algorithm}
    \caption{Parameter Update}
    \label{alg:tspu}
    \KwIn{
        $\boldsymbol \alpha $(Set of parameters $\alpha$), $\boldsymbol \beta $(Set of parameters $\beta$), $\mathcal F$ (the non-dominated set of solutions).
    }
    \KwOut{$\boldsymbol \alpha $(Set of updated parameters $\alpha$), $\boldsymbol \beta $(Set of updated parameters $\beta$).}
    \For{$\boldsymbol f_i$ in $\mathcal F$} {
        \textbf{if} $f_i=1$ \textbf{then} $r_i \gets 1$~\textbf{else} $r_i \gets 0$ \;
        $\alpha_i \gets \alpha_i + r_i$\;
        $\beta_i \gets \beta_i + 1 - r_i$ \;
    }
\end{algorithm}
\begin{algorithm}
    \caption{Evolution Directions Sampling}
    \label{alg: tsov}
    \KwIn{
        $\mathcal D = \{ \boldsymbol d \}$ (A set of directions), $\mathcal S=\{s_i\}$(A set of solutions), $n_d$ (Number of recommended directions), $e$ (Function evaluation for Thompson sampling), $\mathcal{E}$ (An evolutionary optimizer).
    }
    \KwOut{$\mathcal {\hat D} = \{ \boldsymbol d \}$ (A set of sampled directions), $\{\mathcal {S}_i\}$ (Grouped solutions).}
    $\mathcal {\hat D} \gets \emptyset$, $\mathcal {\hat S} \gets \emptyset $, $e' \gets 0$ \; 
    $\{\mathcal D_i \} \gets $ Divide the set $\mathcal D$ into $n_d$ groups randomly\;
    $\{\mathcal S_i \} \gets $ Divide the set $\mathcal S$ into $n_d$ groups randomly\;
    
    \For{$\mathcal D_i$ in $\{\mathcal D_i \}$ and $\mathcal S_i$ in $\{\mathcal S_i \}$} {
      $n \gets $ size of $\mathcal D_i$ \;
      $\boldsymbol \alpha \gets \boldsymbol 1 \in \mathbb R^{n\times 1}$; $\boldsymbol \beta \gets \boldsymbol 1 \in \mathbb R^{n\times 1}$\;
      \While {$e' \leq e$ } {
        \textbf{for} $d_j \in \mathcal D_i$ \textit{and} $\boldsymbol s_j \in \mathcal S_d$ \textbf{do} $\boldsymbol s_j \gets \boldsymbol s_j + \boldsymbol d_j$\;
        $\mathcal F \gets $ \textbf{Fast-Non-Dominated-Sort}($\mathcal S_i$) \;
        $ e' + \vert \mathcal S_i \vert$; // \textit{Record the number of function evaluations used on set} $\mathcal S_i$. \\
        \textbf{Parameter-Update}$(\boldsymbol \alpha, \boldsymbol \beta, \mathcal F)$ \;
        $S_i \gets \mathcal{E}(S_i, \mathcal{F})$ // \textit{Use evolutionary algorithm on $S_i$ with evaluation of $\mathcal F$ to obtain new $S_i$} \;
      }
      \For {$ i \gets 1$ to $n$} {
        Sample the estimated mean reward $\bar \theta_i$ from Beta distribution $P^{Beta}(\alpha_i, \beta_i)$ \;
      }
      $k \gets \arg\max_{i\in \{1,\dots,n\}} \bar \theta_i$\;
      $\mathcal {\hat D} \gets \mathcal {\hat D} \cup \{\boldsymbol d_k \}$\; 
    }
\end{algorithm}
In our approach, we employ the Thompson sampling algorithm to sample and suggest optimal directions. The item $i_j$ represents the direction $\boldsymbol d_j$. Subsequently, the reward $r_{ij}$ for the direction $d_j$ is determined by the quality of the solution $\boldsymbol s_i$ following the application of the evolution algorithm in the direction of $\boldsymbol d_j$. The dominance of the solution is utilized to assess the quality of the solution:
\begin{equation}
\label{equ:reward}
\begin{aligned}
r_{ij} =  \begin{cases}
    1,& \text{$s_i$ evolved by\ } \boldsymbol d_j \text{\ Pareto-dominates\ } \boldsymbol s_i \\
    0,& \text{Otherwise}.
  \end{cases}
\end{aligned}
\end{equation}
\par
Since an evolutionary algorithm is adopted when evolving directions, the fixed reward distribution in Eq.~(\ref{equ: beta}) does not fit the non-stationary nature of the possible solutions. To address this problem, a dynamic Thompson sampling \cite{6147024} strategy is adopted, which is presented in Algorithm~\ref{alg:tspu}. $\alpha$ and $\beta$ of each distribution are initialized to be 1, and $P^{Beta}(\theta_i)$ is initialized to be a uniform distribution over $[0, 1]$.
\par
The Thompson sampling algorithm recommends the direction $\boldsymbol d$. The reward $r_{ij}$ of using direction $\boldsymbol d_j$ by for solution $\boldsymbol s_i$ is calculated by evaluating the new solution $\boldsymbol {\hat s}_i$ evolved with $\boldsymbol d_j$. For each group of solutions $\mathcal S_i$, we will recommend the best direction $\boldsymbol d_k$ from the group of directions $\mathcal D_i$. 
An illustrative example of Thompson sampling on the direction method is given in Fig.~\ref{fig: TSonDir}, and the algorithm for sampling directions is described in Algorithm~\ref{alg: tsov}. Line 2 and Line 3 partition the directions and solutions into $n_d$ groups. \textbf{Fast-Non-dominated-Sort} is the process derived from NSGA-II \cite{996017}. Line 11 records consumed function evaluations.
\subsection{Evolution Directions Fine-tuning}
To search for a set of Pareto-optimal solutions in a very large-scale decision space, a key problem that needs to be solved is the sufficiency of evolution directions. Direction sampling based on Thompson sampling provides us with a small number of potential evolution directions. To address the insufficiency of evolution directions, a method of fine-tuning directions for each solution is proposed in this Section.
\par
The essence of evolution directions fine-tuning is to fine-tune the potential directions obtained by the recommendation, so as to generate a unique and suitable evolution direction for each solution.
\par
For the paired group of solutions $\mathcal S_i$ and group of directions $\mathcal D_d$, we have sampled and recommended direction $\boldsymbol d_k$. To achieve the purpose of fine-tuning for each solution, a direction population is generated based on the recommended direction $\boldsymbol d_k$ and select a set of representative solutions from the group of solutions $\mathcal{S}_i$ to evaluate the corresponding population. 
\par
{The algorithm selects representative solutions with the largest crowding distance~\cite{996017} on the first non-dominated front from grouped solutions $\mathcal S_i$, including boundary solutions.} For each representative solution $\boldsymbol s_d$, the algorithm firstly initializes a set of random direction population $\mathcal D_p =\{\boldsymbol d_i + \boldsymbol d_k\}$ based on recommended directions $\boldsymbol d_k$, the size of $\mathcal D_p$ is $n_{di} = N / n_d$. Second, solution $\boldsymbol s_d$ will be evolved with direction population and evaluated. Afterward, we apply an evolutionary algorithm $\mathcal{E}$ on the direction population and repeat.
\par
\begin{figure}[ht]
    \centering
    \includegraphics[width=1\hsize]{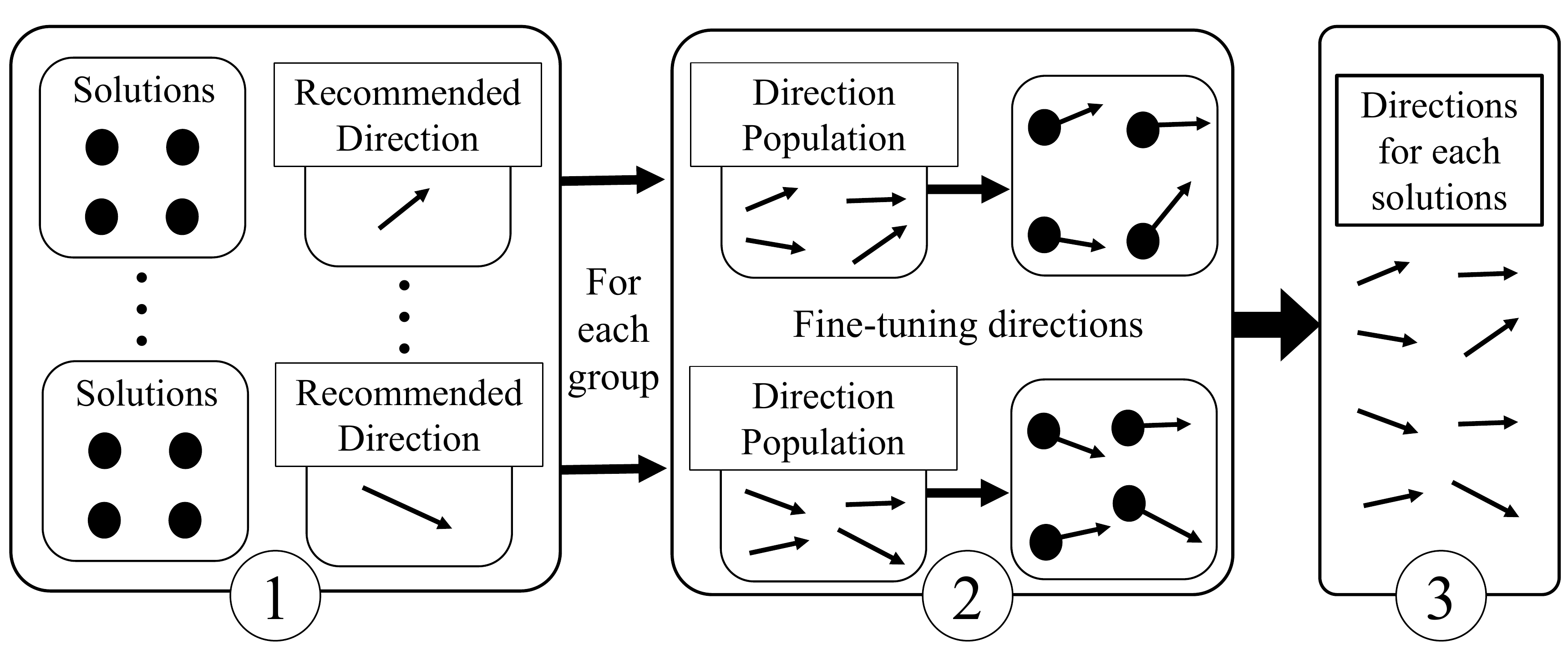}
    \caption{Illustration of evolution directions fine-tuning. 1. For paired group of solutions $\mathcal S_i$ and group of directions $\mathcal D_d$. 2. Initialize a set of direction populations based on the recommended direction, then evolve each solution with direction population and evaluate; 4. Obtain a set of directions for each solution.}
    \label{fig: VelocityOptimization}
\end{figure}
The fine-tuning of evolution directions is performed on each grouped solution $\mathcal S_i$ and recommended direction $\boldsymbol d_k$. Finally, a set of directions is obtained for each solution based on the recommended direction. The visualization of the fine-tuning of the directions is presented in Fig.~\ref{fig: VelocityOptimization} and the process is given in Algorithm~\ref{alg: so}.
\begin{algorithm}
    \caption{Evolution Directions Fine-tuning}
    \label{alg: so}
    \KwIn{
        $\mathcal {\hat D} = \{ \boldsymbol d_k \}$ (Recommended directions), $\{\mathcal S_i \}(\mathcal S_i=\{\boldsymbol s_i\}$)(Grouped solutions), $e$ (Function evaluation for fine-tuning), $\mathcal{E}$ (An evolutionary optimizer).
    }
    \KwOut{$\mathcal D = \{ \boldsymbol d_i \}$ (Fine-tuned directions), $\mathcal {S}$ (Solutions produced during fine-tuning).}
    $N \gets \sum_{i=0}^{i<|\{\mathcal S_i \}|} |\mathcal S_i|$; // \textit{Size of the whole population.}\\
    $n_d \gets |\mathcal {\hat D}|$; // \textit{Size of recommended directions.} \\
    $n_{di} \gets N / n_d$; // \textit{Size for Direction Population.} \\
    $\mathcal D \gets \emptyset$, $e' \gets 0$\;
    \For{$\boldsymbol d_k \in \mathcal {\hat D}$ and $\mathcal S_i$ in $\{\mathcal S_i\}$} {
      $\boldsymbol s_d \gets $ \textbf{Select diversity solutions} $(\mathcal S_i, n_{di})$ \;
      $\mathcal D_p \gets $ Initialize direction population with size $n_{di}$ based on recommended direction $\boldsymbol d_k$\;
      \While {$e' \leq e$} {
        \For{$\boldsymbol d_i \in \mathcal D_p$} {
          $\boldsymbol s_i \gets \boldsymbol s_d + \boldsymbol d_i$;~~$\mathcal S_p \gets \mathcal S_p  \cup \{ \boldsymbol s_i \} $\;
        }
        $\mathcal F \gets $ \textbf{Fast-Non-Dominated-Sort}($\mathcal S_p$) \;
        $ e' + \vert \mathcal S_p \vert$; // \textit{Record the number of function evaluations used on set} $\mathcal S_p$. \\
        $D_p \gets \mathcal{E}(D_p, \mathcal{F})$ // \textit{Use evolutionary algorithm on $\mathcal D_p$ with evaluation of $\mathcal F$ to obtain new $\mathcal D_p$} \;
      }
      \For{$ i \gets 1$ to $n_{di}$} {
        $\mathcal D \gets \mathcal D \cup \{ \boldsymbol d_i \}$; $\mathcal S \gets \mathcal S  \cup \mathcal S_p $\;
      }
    }
\end{algorithm}
\subsection{A Very Large-scale Multiobjective Optimization Framework}
The whole framework of the algorithm is presented in Algorithm~\ref{alg: VMOF} and the initialization of the population and directions is conducted in a random manner. The key to the proposed VMOF lies in performing Thompson sampling for global search and fine-tuning directions for a more granular search. In the optimization step delineated in Line 6, the particle swarm algorithm is adopted due to its inherent compatibility with the sampled and fine-tuned directions. The algorithm initializes the particle swarm with the direction set, denoted as $\mathcal D$. Subsequent to the particle swarm optimization, the direction corresponding to each solution is documented and utilized as the initial direction for the forthcoming iteration of evolution directions sampling.
\begin{algorithm}
    \caption{VMOF}
    \label{alg: VMOF}
    \KwIn{
        $N$ (population size), $n_d$ (recommended directions size), $e$ (Function evaluation for sampling, fine-tuning, and evolving, $\mathcal{E}$ (An evolutionary optimizer).)
    }
    \KwOut{$S$ (A set of solutions optimized by VMOF).}
    $\mathcal S \gets $ Random initialize $N$ solutions \;
    $\mathcal D \gets $ Random initialize $N$ directions \;
    \While {\text{termination criterion is not fulfilled}} {
      $(\mathcal {\hat D}, \{\mathcal S_i \}) \gets $ \textbf{Evolution Directions Sampling} $(\mathcal D, \mathcal S, n_d, e, \mathcal{E})$ \;
      $(\mathcal D, \mathcal S) \gets $ \textbf{Evolution Directions Fine-tuning} $(\mathcal {\hat D}, \{\mathcal S_i \}, e, \mathcal{E})$ \;
      $(\mathcal D, \mathcal S) \gets$ \textbf{Optimizer} $(\mathcal D, \mathcal S, e)$ \;
    }
\end{algorithm}
\section{Analytical Studies}
\label{sec: ana}
This section discusses some possible theoretical foundations to support the design of direction sampling and direction fine-tuning and analyzes the time complexity of the proposed framework.
\subsection{Direction Sampling in Very Large-scale Search Space}
Firstly, the reward of direction $i$ is depicted by the relationship of Pareto-dominance between the old solution and new solution with evolved along the direction. Therefore, rewards, i.e., dominance relationships, are easily represented as binary Consequently, the Beta distribution, which is naturally good for Bernoulli rewards, is adopted to establish distributions on directions.
\par
Secondly, Thompson sampling has been proved to match the asymptotic rate for the cumulative regret in the case of Bernoulli rewards~\cite{10.1007/978-3-642-34106-9_18}, which forms the basis for using Thompson sampling to recommend best solutions. Here we give a problem-independent regret bound for Thompson sampling using Beta priors: 
\begin{align}
    \mathbb{E}[r(T)] \leq \mathcal{O} (\sqrt{ST\ln T})
\end{align}
where $T$ is the time and $S$ is the number of available candidates. Given population size $N$, the number of recommended directions $n_d$, and function evaluations for Thompson sampling $e$, $T$ is the iteration of updating Thompson sampling ($T=e/|S_i|=e/(N/n_d)$, Line 11 of Algorithm \ref{alg: tsov}), and the $S$ is $n_d$, therefore the regret bound for direction sampling is $\mathbb{E}[r] \leq \mathcal{O} (\sqrt{(en_d^2)/N\ln{(en_d/N)}})$.
\subsection{Direction Fine-tuning in Very Large-scale Search Space}
The curse of dimensionality arises when the optimum is situated in a large-scale search space, illustrating the challenges of optimization through exhaustive enumeration on product spaces~\cite{donoho2000}. In the context of MOPs, this can be informally characterized as the requirement of $n(1/\epsilon)^{d}$ function evaluations to obtain $n$ solutions, each approximating the true Pareto optimal set (PS) within an error of $\epsilon$, when optimizing $m$ objectives with $d$ decision variables~\cite{donoho2000}.
\par
The term $n$ signifies that VLSMOPs necessitate a set of solutions, a characteristic derived from their multi-objective nature. The term $(1/\epsilon)^{d}$ is directly associated with the dimension of the problem's decision variables. The expression $n(1/\epsilon)^{d}$ underscores the difficulty of locating a set of PS in an expansive search space, thereby emphasizing the need for fine-tuning the exploration directions for each solution to thoroughly probe the search space.
\subsection{Time Complexity Analysis}
Given an MOP with $d$-dimensional decision variable and $m$ objectives, for the proposed VMOF, the population size is $N$, number of sampled directions is $n_d$, the total function evaluation is $E$, and function evaluation for directions sampling, directions fine-tuning, and optimization algorithm are $e$. Algorithm \ref{alg: tsov} is essentially updating Beta distribution for $N$ directions for $en_d/N$ times. Each update evolves a $d$-dimensional solution with the direction and then evaluates for sorting, so the time complexity of Algorithm \ref{alg: tsov} is $\mathcal{O}(Nd(en_d/N)MN^2)=\mathcal{O}(den_dMN^2)$. Algorithm \ref{alg: so} is fine-tuning $N$ $d$-dimensional directions, and each fine-tuning repeated $e/|S_p|=en_d/N$ times, then the time complexity of \ref{alg: so} is  $\mathcal{O}(den_dMN^2)$, adopted particle swarm optimization has the time complexity with $\mathcal{O}(MN^2de)$, therefore the total time complexity of VMOF is $E/e [\mathcal{O}(den_dMN^2) + \mathcal{O}(den_dMN^2) + \mathcal{O}(MN^2de)] = \mathcal{O}(dMN^2E)$.
\section{Experimental Studies}
\label{sec:exp}
\subsection{Algorithms Used in Comparison and Test Problems}
The proposed VMOF is compared with several state-of-the-art MOEAs, including NSGA-II\cite{996017}, CCGDE3 \cite{6557903}, WOF \cite{RN89}, LSMOF\cite{RN88}, LMOCSO\cite{8681243}, GMOEA \cite{9082904} and  DGEA \cite{9138459}. All the compared algorithms are implemented on PlatEMO \cite{RN92}. NSGA-II, as the baseline, is an MOEA for solving MOPs. CCGDE3 is a typical method for solving the LSMOP by decision variable grouping. WOF and LSMOF are two representative frameworks based on decision space reduction. LMOCSO and LMOEA-DS represent two promising methodologies that leverage novel search strategies, employing competitive swarm optimization and direct sampling, respectively. DGEA, while also utilizing concepts such as sampling and direction vectors, diverges significantly from our approach. DGEA selects a balanced parent population and utilizes these selections to construct direction vectors within decision spaces, thereby facilitating the generation of promising offspring solutions.
\par
The benchmark LSMOP \cite{7553457} is widely used for comparing the performance of large-scale MOEAs. 
We also included TREE \cite{8962275} in the experiments as a real-world problem. The problem has dimensions up to 900,000, which constitutes VLSMOPs. In the experiments, the IGD \cite{RN98} and HV \cite{1583625} metrics are utilized to evaluate the convergence and the diversity of compared algorithms. 
\par
For testing the statistical significance of the differences between metric values, each algorithm is run 20 times independently, and we use the Wilcoxon rank-sum test \cite{Haynes2013} to compare the statistic results, we report a difference at a significance level of 0.05. Symbols "+", "-" and "=" indicate the algorithm result is significantly better than, significantly worse than, or statistically not different from the proposed VMOF algorithm.
\subsection{Parameter Settings}
\subsubsection{Decision Variables} The number of decision variables is 100,000, 500,000, and 1,000,000 for VLSMOPs, and 100 to 20,000 for LSMOPs.
\subsubsection{Population Size} \textcolor{black}{Following the setting of population size for bi-objective and tri-objective problems in WOF \cite{RN89}, LSMOF\cite{RN88}, and DGEA \cite{9138459}, the population size $N$ is set to 100 for test benchmarks with two objectives and 105 for test benchmarks with three objectives.}
\subsubsection{Termination Condition}
The maximum number of evaluations, denoted as $E$, is established at 100,000. While this quantity of function evaluations might appear small, it is deemed practical for real-world applications \cite{RN88}. This is primarily due to the economic and/or computational costs that invariably limit the number of function evaluations, particularly in the context of large-scale optimization problems.
\par
For algorithms that necessitate an extensive amount of time to resolve VLSMOP or LSMOP, the function evaluation would not serve as a fair termination condition. We have documented the time expended on solving the 1,000-dimensional LSMOP1 in Table~\ref{tab: pre-test} as preliminary experiments for algorithm selection. Algorithms that require more than 1,000 seconds, which is five times longer than typical methods, are highlighted in bold. For comparative purposes, we have chosen MOEA/DVA as a representative example.
\begin{table}[htbp]

\scriptsize
  \centering
  \caption{Preliminary Experiment for Algorithm Selection\\ Runtime of Large-scale MOEAs for Solving 1000-Dimensional Bi-objective LSMOP 1. (100,000FEs, unit: s)}
    \begin{tabular}{c|c|c|c|c}
    \toprule
    CCGDE3 & MOEA/DVA & LMEA  & WOF   & LSMOF \\
    \midrule
    16.61 & \textbf{1709} & \textbf{1582} & 34.26 & 27.58 \\
    \midrule
    MOEA/PSL & LMOCSO & LMOEA-DS & DGEA  & VMOF \\
    \midrule
    \textbf{4993} & 28.15 & 14.51 & 28.27 & 25.52 \\
    \bottomrule
    \end{tabular}%
  \label{tab: pre-test}%
\end{table}%
\subsubsection{Parameter Settings for Algorithms}
According to \cite{RN92}, parameters of compared algorithms are set according to their original publications. The specific parameters of each algorithm are given in Table~S-1 of the supplementary material\footnote{The supplementary material can be downloaded from the following link: \href{https://www.jianguoyun.com/p/DeP9v_QQrLOVCxjO-MYFIAA}{Link to the supplementary material}\label{foot:sup}}.
\par
For the proposed algorithm, the sampled directions size $n_d$ is set to $N / 4 = 25$. The function evaluation $e$ for directions sampling, directions fine-tuning, and optimization algorithm are all set to $0.05E$. The evolutionary optimizer $\mathcal{E}$ used in Algorithm \ref{alg: tsov} and \ref{alg: so} is NSGA-II\cite{996017}.
\begin{table*}[htbp]
\scriptsize

  \centering
  \caption{IGD Metric Obtained by Compared Algorithms Over Bi-Objective VLSMOP Problems with Dimension 1,000,000.\\ The Best Result in Each Row is Highlighted in Bold.}
    \begin{tabular}{cccccccccccc}
    \toprule
    Problem & D     & NSGA-II & CCGDE3 & MOEA/DVA & WOF   & LSMOF & LMOCSO & LMOEA-DS & DGEA  & VMOF  &  \\
    \midrule
    LSMOP1 & 1,000,000 & 1.17e+01- & 6.39e+00- & 1.18e+01- & 3.11e-01- & 3.48e-01- & 1.80e+00- & 3.33e-01- & 3.34e-01- & \textbf{1.59e-01} & \textbf{48.83\%} \\
    \midrule
    LSMOP2 & 1,000,000 & 4.97e-03= & 5.00e-03= & 7.13e-03= & 3.72e-03= & 4.54e-03= & 3.82e-03= & 4.11e-03= & \textbf{3.59e-03=} & 4.30E-03 & -19.91\% \\
    \midrule
    LSMOP3 & 1,000,000 & 3.96e+01- & 3.42e+01- & 4.14e+02- & 6.36e+02- & 1.58e+00= & 2.20e+03- & 1.58e+00= & 1.36e+01- & \textbf{1.57e+00} & \textbf{0.31\%} \\
    \midrule
    LSMOP4 & 1,000,000 & 7.41e-03= & 7.04e-03= & 9.97e-03= & 3.73e-03= & 4.55e-03= & 5.32e-03= & 3.82e-03= & \textbf{3.61e-03=} & 4.41E-03 & -22.23\% \\
    \midrule
    LSMOP5 & 1,000,000 & 2.50e+01- & 1.40e+01- & 2.49e+01- & 7.42e-01- & 7.42e-01- & 3.70e+00- & 7.42e-01- & 7.46e-01- & \textbf{6.84e-01} & \textbf{7.85\%} \\
    \midrule
    LSMOP6 & 1,000,000 & 7.42e-01- & 7.42e-01- & 2.49e+03- & 3.69e-01- & 3.06e-01- & 1.12e+03- & 3.06e-01- & 7.42e-01- & \textbf{1.48e-01} & \textbf{51.56\%} \\
    \midrule
    LSMOP7 & 1,000,000 & 9.21e+04- & 3.74e+04- & 9.57e+04- & 1.52e+00= & 1.52e+00= & 2.29e+03- & 1.52e+00= & 1.52e+00= & \textbf{1.52e+00} & \textbf{0.01\%} \\
    \midrule
    LSMOP8 & 1,000,000 & 2.08e+01- & 1.19e+01- & 2.14e+01- & 2.93e-01- & 7.42e-01- & 3.21e+00- & 7.42e-01- & 7.02e-01- & \textbf{1.55e-01} & \textbf{46.99\%} \\
    \midrule
    LSMOP9 & 1,000,000 & 6.07e+01- & 3.05e+01- & 6.40e+01- & 6.40e+01- & 8.02e-01- & 1.68e+01- & 5.29e-01= & 2.19e+01- & \textbf{5.24e-01} & \textbf{0.95\%} \\
    \midrule
    \multicolumn{2}{c}{(+/-/=)} & {0/7/2} & {0/7/2} & {0/7/2} & {0/6/3} & {0/5/4} & {0/7/2} & {0/4/5} & {0/6/3} &       & 7 \\
    \bottomrule
    \end{tabular}%
  \label{tab:lsmop-igd-million}%
\end{table*}%
\begin{table*}[htbp]
\scriptsize

  \centering
  \caption{HV Metric Obtained by Compared Algorithms Over Bi-Objective VLSMOP Problems with Dimension 1,000,000.\\The Best Result in Each Row is Highlighted in Bold.}
    \begin{tabular}{cccccccccccc}
    \toprule
    Problem & D     & NSGA-II & CCGDE3 & MOEA/DVA & WOF   & LSMOF & LMOCSO & LMOEA-DS & DGEA  & VMOF  &  \\
    \midrule
    LSMOP1 & 1,000,000 & 0.00e+00- & 0.00e+00- & 0.00e+00- & 2.49e-01- & 2.29e-01- & 0.00e+00- & 2.27e-01- & 2.34e-01- & \textbf{4.04e-01} & \textbf{61.94\%} \\
    \midrule
    LSMOP2 & 1,000,000 & 5.80e-01= & 5.80e-01= & 5.75e-01= & 5.82e-01= & 5.81e-01= & 5.82e-01= & 5.81e-01= & \textbf{5.82e-01=} & 5.82E-01 & -0.13\% \\
    \midrule
    LSMOP3 & 1,000,000 & {0.00e+00=} & {0.00e+00=} & {0.00e+00=} & {0.00e+00=} & {0.00e+00=} & {0.00e+00=} & {0.00e+00=} & {0.00e+00=} & {0.00e+00} & 0.00\% \\
    \midrule
    LSMOP4 & 1,000,000 & 5.76e-01= & 5.76e-01= & 5.71e-01= & 5.82e-01= & 5.81e-01= & 5.79e-01= & 5.81e-01= & \textbf{5.82e-01=} & 5.81E-01 & -0.17\% \\
    \midrule
    LSMOP5 & 1,000,000 & 0.00e+00- & 0.00e+00- & 0.00e+00- & \textbf{9.09e-02=} & \textbf{9.09e-02=} & 0.00e+00- & 9.09e-02= & 8.13e-02= & \textbf{9.09e-02} & 0.00\% \\
    \midrule
    LSMOP6 & 1,000,000 & 9.08e-02- & 9.08e-02- & 0.00e+00- & 9.62e-02- & 1.08e-01- & 0.00e+00- & 1.08e-01- & 9.09e-02- & \textbf{1.68e-01} & \textbf{55.96\%} \\
    \midrule
    LSMOP7 & 1,000,000 & {0.00e+00=} & {0.00e+00=} & {0.00e+00=} & {0.00e+00=} & {0.00e+00=} & {0.00e+00=} & {0.00e+00=} & {0.00e+00=} & {0.00e+00} & 0.00\% \\
    \midrule
    LSMOP8 & 1,000,000 & 0.00e+00- & 0.00e+00- & 0.00e+00- & 1.19e-01- & 9.09e-02- & 0.00e+00- & 9.09e-02- & 4.91e-03- & \textbf{2.11e-01} & \textbf{76.97\%} \\
    \midrule
    LSMOP9 & 1,000,000 & 0.00e+00- & 0.00e+00- & 0.00e+00- & 0.00e+00- & 9.19e-02- & 0.00e+00- & \textbf{1.53e-01=} & 0.00e+00- & 1.51E-01 & -0.95\% \\
    \midrule
    \multicolumn{2}{c}{(+/-/=)} & {0/5/4} & {0/5/4} & {0/5/4} & {0/4/5} & {0/4/5} & {0/5/4} & {0/3/6} & {0/4/5} &       & 6 \\
    \bottomrule
    \end{tabular}%
  \label{tab:million_hv}%
\end{table*}%

\begin{table*}[htbp]
\scriptsize

  \centering
  \caption{IGD Metric Obtained by Compared Algorithms Over Bi-Objective VLSMOP Problems with Dimension 100,000, and 500,000.\\The Best Result in Each Row is Highlighted in Bold.}
    \begin{tabular}{cccccccccccc}
    \toprule
    Problem & D     & NSGA-II & CCGDE3 & MOEA/DVA & WOF   & LSMOF & LMOCSO & LMOEA-DS & DGEA  & VMOF  & ROC (\%) \\
    \midrule
    \multirow{2}[2]{*}{LSMOP1} & 100,000 & 1.28e+01- & 1.07e+01- & 1.17e+01- & 4.37e-01- & 7.49e-01- & 1.67e+00- & 3.20e-01- & 4.33e-01- & \textbf{2.00e-01} & \textbf{53.81\%} \\
          & 500,000 & 1.55e+01- & 1.36e+01- & 1.17e+01- & 4.83e-01- & 1.10e+00- & 2.43e+00- & 3.41e-01- & 9.20e-01- & \textbf{1.82e-01} & \textbf{62.32\%} \\
    \midrule
    \multirow{2}[2]{*}{LSMOP2} & 100,000 & 6.22e-02= & 4.98e-02= & 7.24e-03= & \textbf{3.82e-02=} & 4.99e-02= & 3.88e-02= & 4.26e-03= & 3.85e-02= & 5.02E-02 & -31.41\% \\
          & 500,000 & 8.89e-02= & 7.60e-02= & 7.09e-03= & 5.83e-02= & 7.51e-02= & 5.88e-02= & 6.75e-03= & 5.80e-02= & \textbf{5.05e-02} & \textbf{12.93\%} \\
    \midrule
    \multirow{2}[2]{*}{LSMOP3} & 100,000 & 3.88e+01- & 1.84e+01- & 3.21e+02- & \textbf{8.61e-01=} & \textbf{8.61e-01=} & 3.90e+01- & 1.58e+00= & 8.62e-01= & \textbf{8.61e-01} & 0.00\% \\
          & 500,000 & 3.31e+01- & 2.90e+01- & 1.94e+03- & 1.29e+00- & 1.29e+00- & 2.55e+01- & 2.76e+03- & 1.29e+00- & \textbf{8.61e-01} & \textbf{33.26\%} \\
    \midrule
    \multirow{2}[2]{*}{LSMOP4} & 100,000 & 5.62e-02= & 5.99e-02= & 9.63e-03= & 5.82e-02= & 5.24e-02= & 5.83e-02= & 4.22e-03= & 5.82e-02= & \textbf{3.30e-02} & \textbf{37.02\%} \\
          & 500,000 & 8.33e-02= & 7.85e-02= & 1.07e-02= & 6.17e-02= & 7.85e-02= & 6.20e-02= & 1.06e-02= & 6.04e-02= & \textbf{5.72e-02} & \textbf{5.30\%} \\
    \midrule
    \multirow{2}[2]{*}{LSMOP5} & 100,000 & 2.13e+01- & 1.71e+01- & 2.50e+01- & 5.89e-01= & 9.46e-01- & 3.44e+00- & 7.42e-01- & 5.57e-01- & \textbf{5.01e-01} & \textbf{10.05\%} \\
          & 500,000 & 3.13e+01- & 2.52e+01- & 2.49e+01- & 6.71e-01- & 1.42e+00- & 5.34e+00- & 2.49e+01- & 2.73e+00- & \textbf{3.60e-01} & \textbf{46.35\%} \\
    \midrule
    \multirow{2}[2]{*}{LSMOP6} & 100,000 & 6.75e+04- & 4.88e+04- & 1.59e+03- & 1.71e+00- & 7.81e-01- & 1.39e+03- & 3.06e-01- & 5.07e+02- & \textbf{7.05e-01} & \textbf{9.73\%} \\
          & 500,000 & 5.75e+04- & 2.98e+04- & 5.56e+03- & 1.34e+00- & 8.47e-01- & 4.44e+02- & 2.73e+03- & 8.36e-01- & \textbf{7.05e-01} & \textbf{15.67\%} \\
    \midrule
    \multirow{2}[2]{*}{LSMOP7} & 100,000 & 9.47e-01- & 9.47e-01- & 9.63e+04- & \textbf{4.57e-01+} & 9.13e-01- & 9.47e-01- & 1.52e+00= & 9.13e-01- & 8.34E-01 & -82.49\% \\
          & 500,000 & 9.46e-01- & 9.46e-01- & 9.62e+04- & 6.69e-01- & 9.13e-01- & 9.46e-01- & 9.55e+04- & 1.32e+00- & \textbf{6.33e-01} & \textbf{5.38\%} \\
    \midrule
    \multirow{2}[2]{*}{LSMOP8} & 100,000 & 9.49e-01- & 9.49e-01- & 2.13e+01- & 1.73e-01- & 2.32e-01- & 9.48e-01- & 7.42e-01- & 9.46e-01- & \textbf{8.01e-02} & \textbf{53.70\%} \\
          & 500,000 & 9.49e-01- & 9.49e-01- & 2.14e+01- & 9.75e-02= & 2.16e-01- & 9.49e-01- & 2.14e+01- & 1.35e+00- & \textbf{7.46e-02} & \textbf{23.49\%} \\
    \midrule
    \multirow{2}[2]{*}{LSMOP9} & 100,000 & 1.24e+02- & 9.05e+01- & 6.38e+01- & 1.15e+00- & 1.49e+00- & 6.88e+01- & 6.08e-01+ & 6.17e+01- & \textbf{5.87e-01} & \textbf{48.96\%} \\
          & 500,000 & 1.44e+02- & 8.92e+01- & 6.39e+01- & 1.27e+00- & 1.14e+00- & 6.63e+01- & 6.40e+01- & 6.89e+01- & \textbf{5.86e-01} & \textbf{48.60\%} \\
    \midrule
    \multicolumn{2}{c}{(+/-/=)} & {0/14/4} & {0/14/4} & {0/14/4} & {1/10/7} & {0/13/5} & {0/14/4} & {1/11/6} & {0/13/5} &       &  \\
    \bottomrule
    \end{tabular}%
  \label{tab:huge_igd}%
\end{table*}%
\begin{table*}[htbp]
\scriptsize
  \centering
  \caption{HV Metric Obtained by Compared Algorithms Over Bi-Objective VLSMOP Problems with Dimension 100,000, and 500,000.\\The Best Result in Each Row is Highlighted in Bold.}
    \begin{tabular}{cccclccclccc}
    \toprule
    Problem & D     & NSGA-II & CCGDE3 & MOEA/DVA & WOF   & LSMOF & LMOCSO & LMOEA-DS & DGEA  & VMOF  & ROC (\%) \\
    \midrule
    \multirow{2}[2]{*}{LSMOP1} & 100,000 & 0.00e+00- & 0.00e+00- & 0.00e+00- & 2.73e-01- & 9.09e-02- & 0.00e+00- & 2.42e-01- & 3.07e-01- & \textbf{5.83e-01} & \multicolumn{1}{r}{\textbf{89.90\%}} \\
          & 500,000 & 0.00e+00- & 0.00e+00- & 0.00e+00- & 3.97e-01- & 9.09e-02- & 0.00e+00- & 2.16e-01- & 1.84e-01- & \textbf{6.26e-01} & \multicolumn{1}{r}{\textbf{57.68\%}} \\
    \midrule
    \multirow{2}[2]{*}{LSMOP2} & 100,000 & 8.21e-01= & 8.28e-01= & 5.76e-01= & \textbf{8.43e-01=} & 8.26e-01= & 8.41e-01= & 5.81e-01= & 8.42e-01= & 8.26E-01 & \multicolumn{1}{r}{-2.02\%} \\
          & 500,000 & 8.18e-01= & 8.20e-01= & 5.74e-01= & 8.40e-01= & 8.30e-01= & 8.40e-01= & 5.77e-01= & \textbf{8.41e-01=} & 8.27E-01 & \multicolumn{1}{r}{-1.66\%} \\
    \midrule
    \multirow{2}[2]{*}{LSMOP3} & 100,000 & 0.00e+00- & 0.00e+00- & \multicolumn{1}{c}{0.00e+00-} & \textbf{9.09e-02=} & \textbf{9.09e-02=} & 0.00e+00- & \multicolumn{1}{c}{0.00e+00-} & 8.98e-02= & \textbf{9.09e-02} & \multicolumn{1}{r}{0.00\%} \\
          & 500,000 & 0.00e+00- & 0.00e+00- & \multicolumn{1}{c}{0.00e+00-} & \textbf{9.09e-02=} & \textbf{9.09e-02=} & 0.00e+00- & \multicolumn{1}{c}{0.00e+00-} & \textbf{9.09e-02=} & \textbf{9.09e-02} & \multicolumn{1}{r}{0.00\%} \\
    \midrule
    \multirow{2}[2]{*}{LSMOP4} & 100,000 & 8.20e-01= & 8.30e-01= & 5.70e-01= & \textbf{8.43e-01=} & 8.31e-01= & 8.43e-01= & 5.81e-01= & 8.43e-01= & 8.26E-01 & \multicolumn{1}{r}{-2.02\%} \\
          & 500,000 & 8.11e-01= & 8.20e-01= & 5.69e-01= & 8.35e-01= & 8.24e-01= & 8.35e-01= & 5.71e-01= & \textbf{8.37e-01=} & 8.17E-01 & \multicolumn{1}{r}{-2.39\%} \\
    \midrule
    \multirow{2}[2]{*}{LSMOP5} & 100,000 & 0.00e+00- & 0.00e+00- & 0.00e+00- & 3.36e-01= & 9.09e-02- & 0.00e+00- & 9.09e-02= & 2.49e-01- & \textbf{3.44e-01} & \multicolumn{1}{r}{\textbf{2.38\%}} \\
          & 500,000 & 0.00e+00- & 0.00e+00- & 0.00e+00- & 3.40e-01= & 9.09e-02- & 0.00e+00- & 0.00e+00- & 0.00e+00- & \textbf{3.44e-01} & \multicolumn{1}{r}{\textbf{1.18\%}} \\
    \midrule
    \multirow{2}[2]{*}{LSMOP6} & 100,000 & 0.00e+00= & 0.00e+00= & 0.00e+00- & 0.00e+00= & \textbf{1.20e-02=} & 0.00e+00= & 1.08e-01= & 0.00e+00= & 0.00E+00 & \multicolumn{1}{r}{-100.00\%} \\
          & 500,000 & 0.00e+00- & 0.00e+00- & 0.00e+00- & 0.00e+00- & 3.12e-03- & 0.00e+00- & 0.00e+00- & 7.00e-03- & \textbf{1.18e-01} & \multicolumn{1}{r}{\textbf{1585.71\%}} \\
    \midrule
    \multirow{2}[2]{*}{LSMOP7} & 100,000 & 8.85e-02= & 8.86e-02= & \multicolumn{1}{c}{0.00e+00-} & \textbf{1.58e-01+} & 8.93e-02= & 8.88e-02= & \multicolumn{1}{c}{0.00e+00-} & 9.01e-02= & 9.01E-02 & \multicolumn{1}{r}{-42.97\%} \\
          & 500,000 & 9.04e-02= & 9.04e-02= & \multicolumn{1}{c}{0.00e+00-} & \textbf{9.08e-02=} & 9.05e-02= & 9.05e-02= & \multicolumn{1}{c}{0.00e+00-} & 8.14e-02= & 9.07E-02 & \multicolumn{1}{r}{-0.11\%} \\
    \midrule
    \multirow{2}[2]{*}{LSMOP8} & 100,000 & 8.44e-02- & 8.50e-02- & 0.00e+00- & 4.12e-01- & 3.83e-01- & 8.68e-02- & 9.09e-02- & 9.04e-02- & \textbf{4.95e-01} & \multicolumn{1}{r}{\textbf{20.15\%}} \\
          & 500,000 & 8.44e-02- & 8.53e-02- & 0.00e+00- & 4.60e-01= & 4.00e-01- & 8.58e-02- & 0.00e+00- & 8.90e-02- & \textbf{5.00e-01} & \multicolumn{1}{r}{\textbf{8.70\%}} \\
    \midrule
    \multirow{2}[2]{*}{LSMOP9} & 100,000 & 0.00e+00- & 0.00e+00- & 0.00e+00- & 1.47e-01= & 9.59e-02- & 0.00e+00- & 1.30e-01= & 0.00e+00- & \textbf{1.92e-01} & \multicolumn{1}{r}{\textbf{30.61\%}} \\
          & 500,000 & 0.00e+00- & 0.00e+00- & 0.00e+00- & 1.29e-01- & 1.48e-01= & 0.00e+00- & 0.00e+00- & 0.00e+00- & \textbf{1.92e-01} & \multicolumn{1}{r}{\textbf{29.73\%}} \\
    \midrule
    \multicolumn{2}{c}{(+/-/=)} & {0/11/7} & {0/15/3} & {0/10/8} & {1/5/12} & {0/8/10} & {0/11/7} & {0/11/7} & {0/9/9} &       &  \\
    \bottomrule
    \end{tabular}%
  \label{tab:huge_hv}%
\end{table*}%

\subsection{Performance on VLSMOPs}
In order to test the performance of our proposed algorithm for solving VLSMOPs, we set the decision variables to 1,000,000, and present the IGD values and HV values in Tables~\ref{tab:lsmop-igd-million} and~\ref{tab:million_hv}. We then present the results of IGD and HV with 100,000 and 500,000 decision variables in Table~\ref{tab:huge_igd} and Table~\ref{tab:huge_hv}, respectively. In the last column of each Table, we present the indicator of the rate of change (ROC) between our algorithm and the best results among other algorithms.
\par
On average, the proposed method demonstrates an improvement of 12\% and 19\% over the suboptimal algorithm, as indicated by the IGD values in Table~\ref{tab:lsmop-igd-million} and Table~\ref{tab:huge_igd}. This improvement is particularly noticeable in the LSMOP 8 and LSMOP 9 problems. Given the unique characteristics of these two problems, we conjecture that as the number of decision variables increases to 100,000 or more, the complexity of the Pareto front will increase correspondingly, and the VLSMOP with a mixed fitness landscape is especially complex. Despite these challenges, our algorithm can effectively identify the optimal direction within a reasonable number of function evaluations in such a large-scale search space. This capability is of considerable practical significance for solving such problems.
\par
From Table~\ref{tab:lsmop-igd-million}, it can be observed that our algorithm achieves significant improvement in solving 1,000,000 dimensional VLSMOPs. In LSMOP2 and LSMOP4, the IGD values of our algorithm are slightly lower, but there is no significant difference with the optimal results.
\par
NSGA-II is a classic algorithm, but this algorithm is less effective when dealing with very large-scale search spaces. WOF only has slightly better performance than VMOF on tri-objective LSMOP7. Both LMOEA-DS and DGEA, which also incorporate sampling, are capable of addressing LSMOPs. However, their overall performance falls short when compared to our proposed algorithm. LMOCSO is indeed an effective particle swarm algorithm for addressing LSMOPs, as it searches for directions using competitive particles. However, in higher dimensions, these directions tend to be inefficient without the application of direction sampling and fine-tuning. Consequently, the performance of LMOCSO remains less than optimal.
\par
We further evaluated the performance of our proposed algorithm using the VLSMOP* benchmark~\cite{10254122}, a variant of the original LSMOP benchmark that incorporates a translation transformation. The empirical results, presented in Tables S-II and S-III of the supplementary materials~\textsuperscript{\ref{foot:sup}}, demonstrate that our proposed VMOF consistently outperforms existing MOEAs.
\par
\subsection{Performance on LSMOPs}
In this section, we conducted experiments on bi-objective and tri-objective LSMOPs with decision variables ranging from 100 to 20,000. The IGD and HV values obtained by the algorithms are detailed in Tables S-IV to S-XI of the supplementary material~\textsuperscript{\ref{foot:sup}}.
\par
Overall, VMOF outperformed the eight contemporary state-of-the-art algorithms. For instance, our proposed VMOF achieved the best results in 19 out of 36 test instances with decision variables ranging from 100 to 1,000 (refer to Table S-IV of the supplementary material). VMOF demonstrated significant advantages over LSMOP2, LSMOP5, LSMOP6, and LSMOP8. Even when the dimension extended to 20,000, our algorithm remained significantly superior to all compared algorithms. While competitive in lower dimensions (from 100 to 1,000), our algorithm also excelled in higher dimensions. In Table~S-VII of the supplementary material, our proposed VMOF achieved the best results in 21 out of 36 bi-objective LSMOP test instances. Only WOF had two results that were significantly better than our proposed method, while the other algorithms' best results were not statistically superior under the Wilcoxon rank-sum test.
\par
When comparing dimensions 100 to 1,000 and 5,000 to 20,000, the performance of WOF, LMOCSO, and DGEA all decreased. WOF had more results that were significantly worse than VMOF (12 to 16), and LMOCSO and DGEA had fewer better results. However, our proposed method VMOF had more best results (19 to 21). As the number of decision variables increased, the number of best results obtained by our algorithm also increased, and a pattern was also observed in solving tri-objective LSMOPs (refer to the supplementary material).
\par
Our proposed method VMOF also demonstrated superior results on the HV metric for LSMOPs. Specifically, our method achieved the best results in 16 and 19 bi-objective and tri-objective optimization problems (refer to Tables~S-VI and S-VII), respectively. In summary, our algorithm also achieved commendable results in terms of convergence and diversity for solving LSMOPs. Thus, we infer that the strategy of sampling and recommending directions for solutions is effective in solving both large-scale and very large-scale MOPs.
\subsection{Performance on Real-world VLSMOPs}
One motivation of the proposed research is to solve very large-scale MOPs for real-world applications. To verify the performance of our algorithm on real problems, we test it on the TREE dataset, and the experimental results are shown in Table~\ref{tab:tree-igd}. It can be seen that the proposed algorithm still has a significant advantage in solving a real-world VLSMOP.
\par
The TREE problem mainly describes the voltage transformers' ratio error (RE) estimation with multiple constraints and very large-scale decision variables. In the obtained results for TREE 1-5, it becomes evident that the proposed algorithm consistently outperforms the compared algorithms across a range of dimensions from 100,000 to 900,000. However, for TREE6, none of the compared algorithms produced feasible solutions within the specified termination conditions. This outcome can be attributed to the additional constraints present in TREE 6, which introduces a higher level of complexity. In general, the results affirm the effectiveness of our algorithm in solving real-world VLSMOPs, highlighting its superiority.
\par
\begin{figure}[ht]
    \centering
    \subfloat[DGEA solves TREE1.]{
        \includegraphics[width=0.44\hsize]{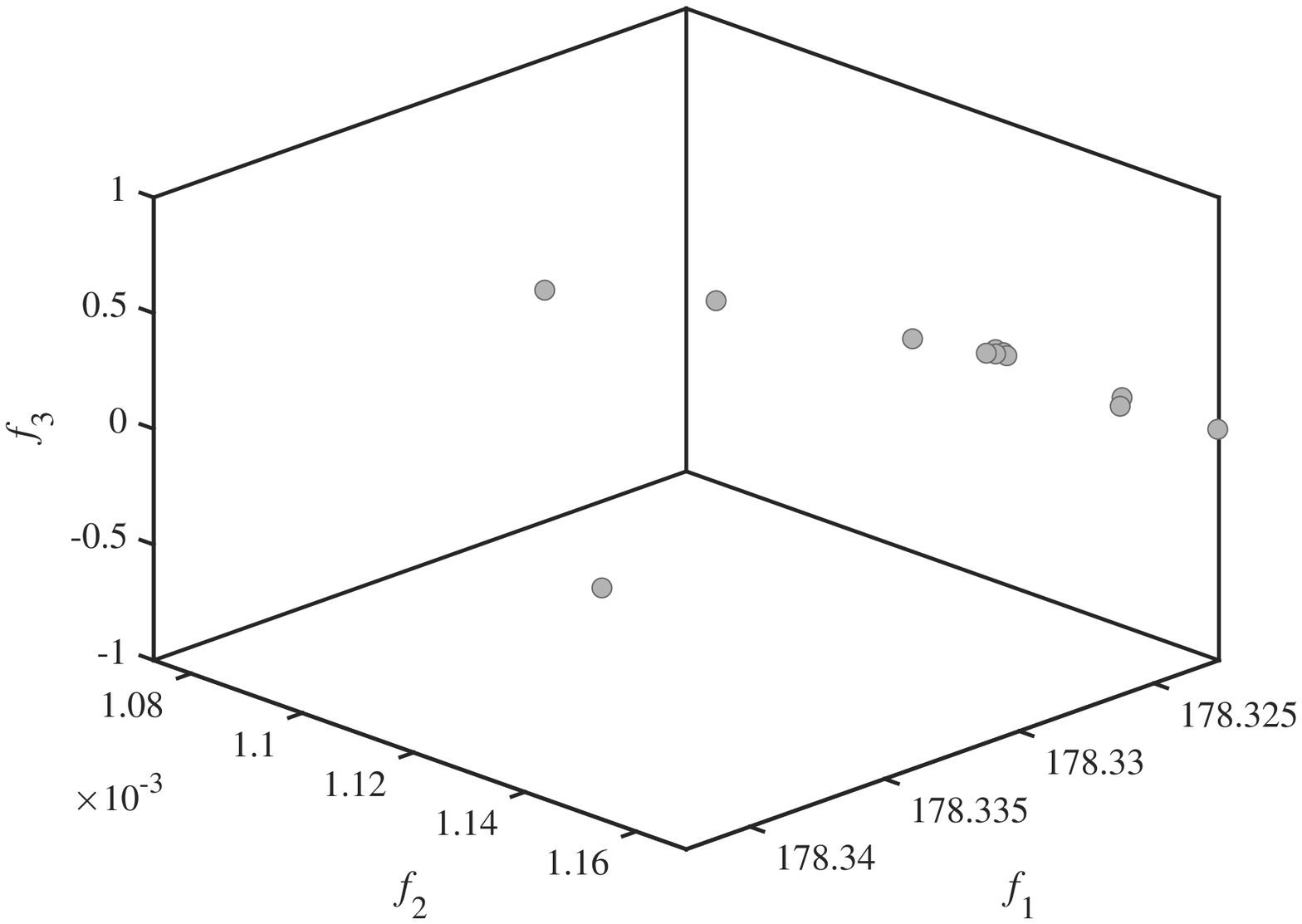}
    }
    \subfloat[VMOF solves TREE1.]{
        \includegraphics[width=0.44\hsize]{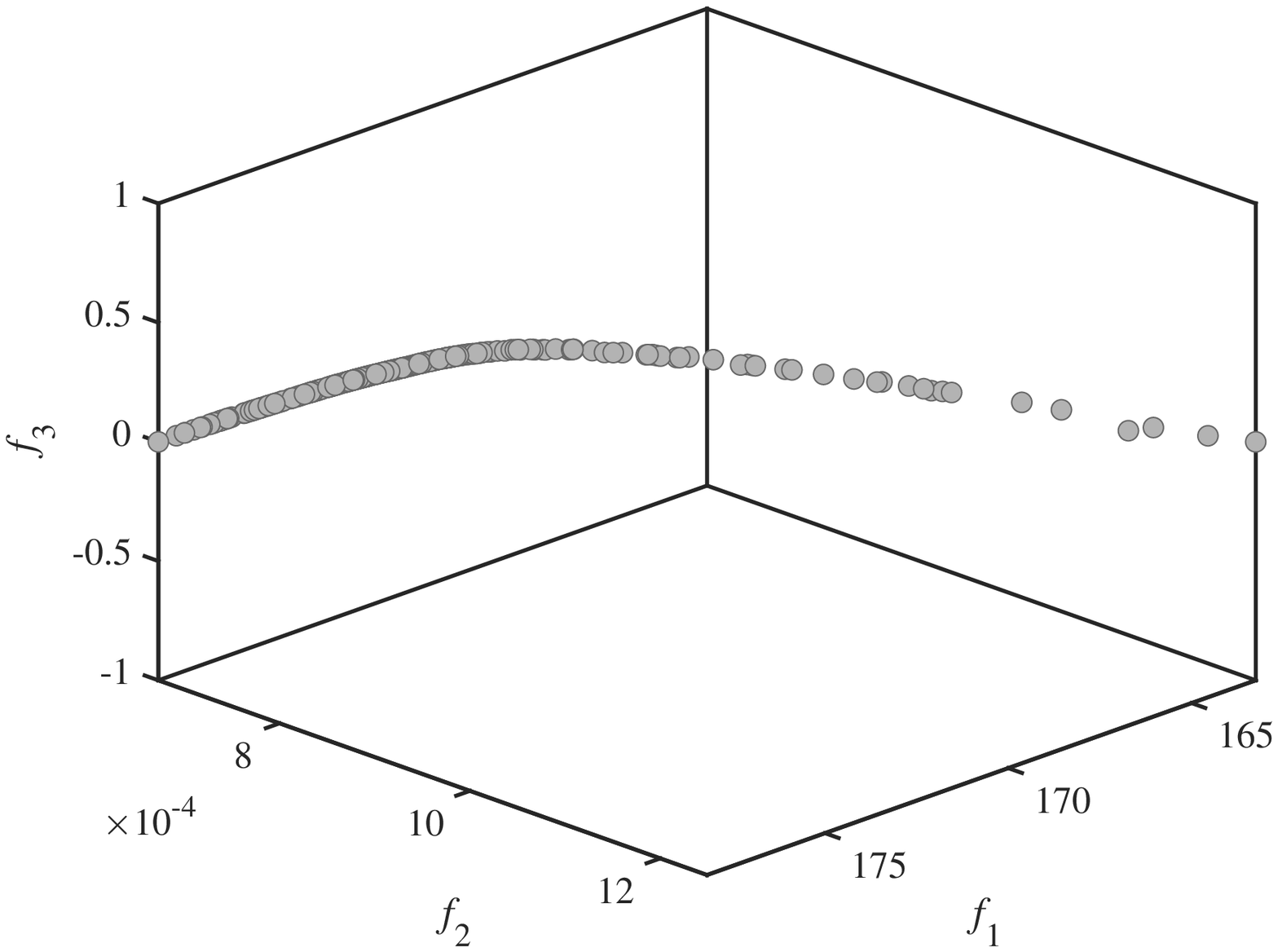}
    }
    \caption{Nondominated solutions obtained by compared methods on TREE with 100,000 decision variables. Please refer to the supplementary materials for other algorithms.}
    \label{fig: pf}
\end{figure}

Besides, nondominated solutions obtained by the proposed VMOF and compared methods on TREE with 100,000 decision variables are presented in Fig.~\ref{fig: pf}, and the figure shows that the set of solutions obtained by the proposed framework has better convergence and diversity.
\begin{table}[htbp]
\scriptsize
  \centering
  \caption{IGD Metric Obtained by Compared Algorithms Over TREE problems.}
  \begin{threeparttable}
  
    \begin{tabular}{ccccccc}
    \toprule
    Problem & D     & WOF   & LSMOF & DGEA  & VMOF & ROC \\
    \midrule
    TREE1 & 100,000 & 9.71e+2- & 9.55e+2- & 9.62e+2- & \textbf{9.53e+2} & 0.20\% \\
    \midrule
    TREE2 & 360,000 & 4.39e+3- & 4.39e+3- & 4.40e+3- & \textbf{4.36e+3} & 0.80\% \\
    \midrule
    TREE3 & 96,000 & 9.60e+2- & 8.72e+2- & 9.60e+2- & \textbf{5.08e+2} & 41.70\% \\
    \midrule
    TREE4 & 120,000 & 1.20e+3- & 1.20e+3- & 1.20e+3- & \textbf{1.05e+3} & 12.81\% \\
    \midrule
    TREE5 & 900,000 & 9.28e+3- & 8.90e+3- & 9.20e+3- & \textbf{8.16e+3} & 9.06\% \\
    \midrule
    TREE6 & 192,000 & -     & -     & -     & -     & - \\
    \midrule
    (+/-/=) &       & {0/5/1} & {0/5/1} & {0/5/1} &       &  \\
    \bottomrule
    \end{tabular}%
    \begin{tablenotes}
        \footnotesize
        \item The best result in each row is highlighted in bold.
        \item Results that took more than 50,000s are represented by '-'.
    \end{tablenotes}
  \end{threeparttable}
  \label{tab:tree-igd}%
\end{table}%
\subsection{Convergence Analysis}
\begin{figure}[ht]
    \centering
    \subfloat[1,000 decision variables.]{
        \includegraphics[width=0.44\hsize]{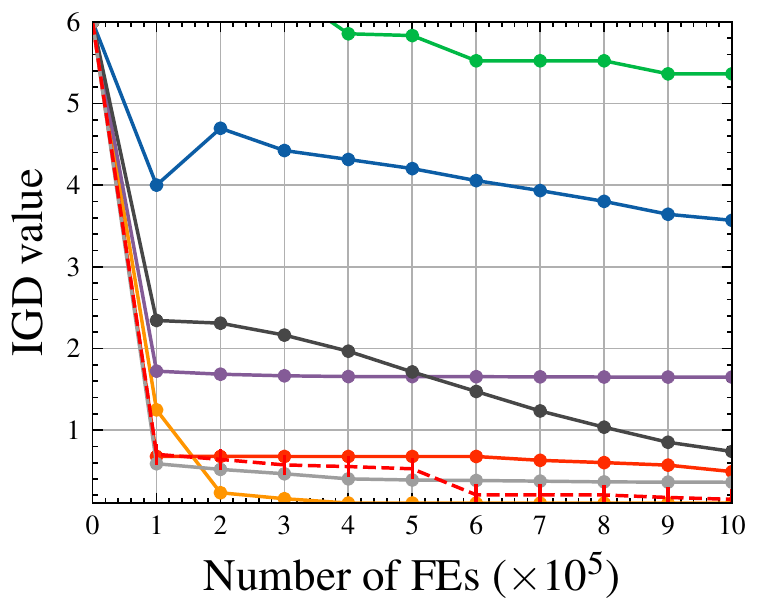}
    }
    \subfloat[100,000 decision variables.]{
        \includegraphics[width=0.44\hsize]{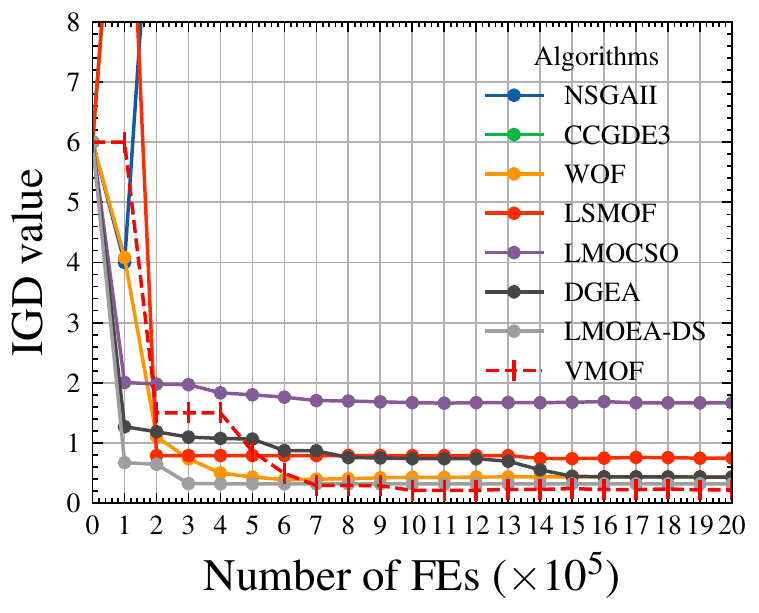}
    }
    \caption{Convergence profiles of all compared algorithms on bi-objective LSMOP1 with 1,000 and 100,000 decision variables, respectively.}
    \label{fig:con_both}
\end{figure}
\par
The variations of IGD values achieved by all compared algorithms on bi-objective LSMOP1 with 1,000 and 100,000 decision variables are presented in Figs.~\ref{fig:con_both}.~(a) and (b), respectively. Due to the large differences between the results of some algorithms, in order to better show the gap between the better algorithms, the figure has been scaled. Consequently, the results of some algorithms in some figures are omitted.
\par
It can be seen from the figures that most algorithms can solve LSMOP1 with 1000 decision variables, but when the problem is expanded to 100,000 dimensions, many algorithms cannot solve it. Specifically, NSGA-II and CCGDE3 fail to converge, whereas LSMOF, WOF, and DGEA have the ability to quickly search for a local optimum within small function evaluations, but fail to converge later. Only our proposed algorithm outputs a good result within the given limit of function evaluations.
\subsection{Time complexity Analysis}
\begin{figure}[ht]
    \centering
    \subfloat[The dimension of the decision variable varies from 100 to 20,000.]{
        \includegraphics[width=0.43\hsize]{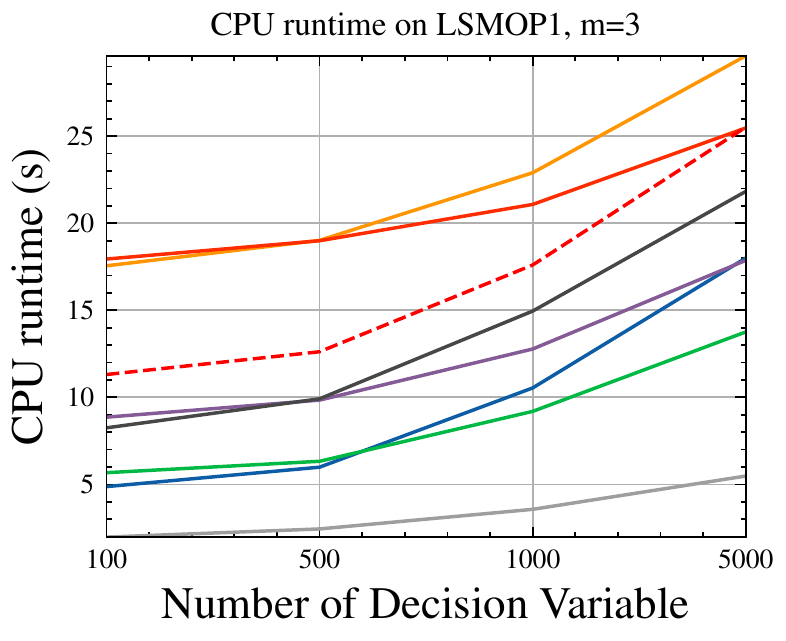}
    }
    \hspace{3mm}
    \subfloat[The dimension of the decision variable varies from 100 to 1,000,000.]{
        \includegraphics[width=0.43\hsize]{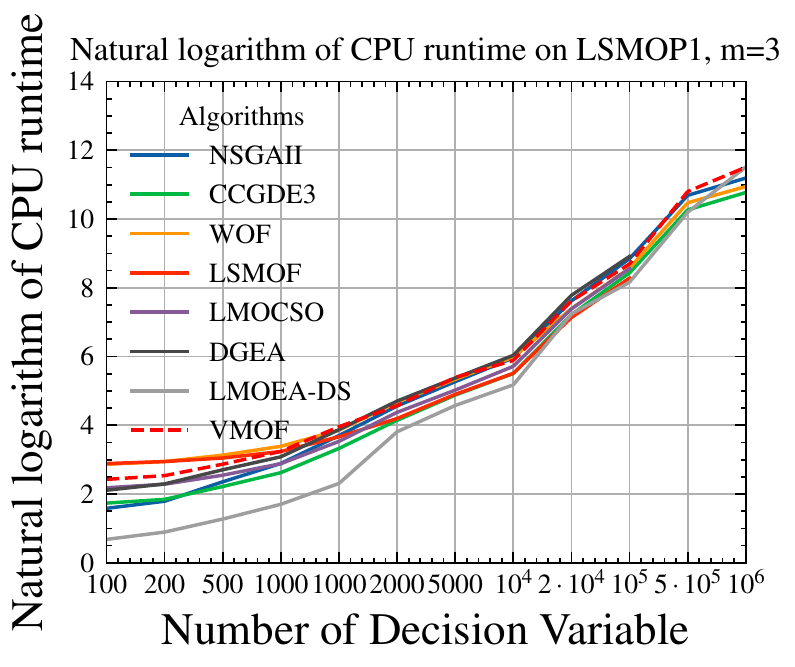}
    }
    \caption{The running time of all compared algorithms in solving bi-objective LSMOP1, where FE=100,000. Please refer to Figs. S-1 to S-9 of the supplementary materials for all results.}
    \label{fig:time_both}
\end{figure}
As an algorithm for solving VLSMOPs and LSMOPs, the time complexity is an important evaluation index. The time complexity of compared algorithms is presented in Figs.~\ref{fig:time_both}.~(a) and (b). The dimensions of the decision variables are from 100 to 5,000 and 100 to 1,000,000, respectively. To better illustrate the results, Figs.~\ref{fig:time_both}.~(b) plots the natural logarithm of the CPU running time from 100 to 1,000,000 dimensions.
\par
Firstly, it can be seen from Figs.~\ref{fig:time_both}.~(a) and (b) that although the time complexity of our algorithm is not the lowest, nevertheless its complexity is not high. Thus, the impact of dimensionality is relatively limited, and our algorithm belongs to the first tier of all running algorithms. Further, the rate with which the time complexity increases with the dimension of decision variables is close to most algorithms. Secondly, as can be seen from Fig.~\ref{fig:time_both}.~(b), the time complexity of all algorithms increases exponentially from 20,000 dimensions to 100,000 dimensions, which simply illustrates that dealing with such large dimensions is a very challenging problem.
\subsection{Ablation Study}

\begin{table}[htbp]
\scriptsize
  \centering
  \caption{Value of IGD Metric Over Tri-Objective VLSMOPs and LSMOPs Obtained by Ablation Study}
  \begin{threeparttable}
    \begin{tabular}{crccc}
    \toprule
    Problem & \multicolumn{1}{c}{D}     & {W/O DR} & {W/O DF} & VMOF \\
    \midrule
    \multirow{4}[2]{*}{LSMOP1} & 10,000 & \textbf{1.96e-01+} & 8.61e-01- & 2.50e-01 \\
          & 20,000 & \textbf{1.37e-01=} & 8.56e-01- & 1.67e-01 \\
          & 100,000 & 2.40e-01= & 8.55e-01- & \textbf{2.00e-01} \\
          & 500,000 & 1.98e-01= & 8.57e-01- & \textbf{1.82e-01} \\
    \midrule
    \multirow{4}[2]{*}{LSMOP2} & 10,000 & \textbf{8.12e-02=} & 1.52e-01- & 8.76e-02 \\
          & 20,000 & 9.35e-02= & \textbf{5.89e-02=} & 7.81e-02 \\
          & 100,000 & 1.20e-01- & 5.31e-02= & \textbf{5.02e-02} \\
          & 500,000 & 6.32e-02= & 6.83e-02= & \textbf{5.05e-02} \\
    \midrule
    \multirow{4}[2]{*}{LSMOP3} & 10,000 & 8.61e-01= & \textbf{8.61e-01=} & 8.61e-01 \\
          & 20,000 & \textbf{8.61e-01=} & \textbf{8.61e-01=} & \textbf{8.61e-01} \\
          & 100,000 & \textbf{8.61e-01=} & \textbf{8.61e-01=} & \textbf{8.61e-01} \\
          & 500,000 & \textbf{8.58e-01=} & 8.61e-01= & 8.61e-01 \\
    \midrule
    \multirow{4}[2]{*}{LSMOP4} & 10,000 & \textbf{5.91e-02=} & 6.72e-02= & 8.54e-02 \\
          & 20,000 & 7.60e-02= & \textbf{5.45e-02=} & 7.05e-02 \\
          & 100,000 & 6.75e-02= & 8.04e-02= & \textbf{3.30e-02} \\
          & 500,000 & 6.63e-02= & 6.53e-02= & \textbf{5.72e-02} \\
    \midrule
    \multirow{4}[2]{*}{LSMOP5} & 10,000 & \textbf{4.98e-01=} & 9.46e-01- & 5.28e-01 \\
          & 20,000 & \textbf{5.01e-01=} & 9.46e-01- & 5.35e-01 \\
          & 100,000 & 5.29e-01= & 9.46e-01- & \textbf{5.01e-01} \\
          & 500,000 & 5.24e-01- & 9.46e-01- & \textbf{3.60e-01} \\
    \midrule
    \multirow{4}[2]{*}{LSMOP6} & 10,000 & \textbf{7.33e-01+} & 1.70e+00- & 8.29e-01 \\
          & 20,000 & \textbf{7.63e-01+} & 1.71e+00- & 1.33e+00 \\
          & 100,000 & 1.71e+00- & 1.71e+00- & \textbf{7.05e-01} \\
          & 500,000 & 8.11e-01- & 1.71e+00- & \textbf{7.05e-01} \\
    \midrule
    \multirow{4}[2]{*}{LSMOP7} & 10,000 & \textbf{8.35e-01=} & 1.71e+00- & 8.35e-01 \\
          & 20,000 & 8.34e-01- & 9.19e-01- & \textbf{7.63e-01} \\
          & 100,000 & 8.35e-01= & 9.14e-01- & \textbf{8.34e-01} \\
          & 500,000 & 8.34e-01- & 1.71e+00- & \textbf{6.33e-01} \\
    \midrule
    \multirow{4}[2]{*}{LSMOP8} & 10,000 & \textbf{8.03e-02=} & 5.44e-01- & 8.66e-02 \\
          & 20,000 & 9.64e-02= & 5.00e-01- & \textbf{7.19e-02} \\
          & 100,000 & 9.88e-02= & 3.50e-01- & \textbf{8.01e-02} \\
          & 500,000 & 1.12e-01- & 5.97e-01- & \textbf{7.46e-02} \\
    \midrule
    \multirow{4}[2]{*}{LSMOP9} & 10,000 & 1.14e+00- & 1.54e+00- & \textbf{5.88e-01} \\
          & 20,000 & 1.14e+00- & 1.54e+00- & \textbf{5.87e-01} \\
          & 100,000 & 1.15e+00- & 1.54e+00- & \textbf{5.87e-01} \\
          & 500,000 & 9.34e-01- & 1.54e+00- & \textbf{5.86e-01} \\
    \midrule
    \multicolumn{2}{c}{(+/-/=)					} & {3/11/22} & {0/25/11} &  \\
    \bottomrule
    \end{tabular}%
    \begin{tablenotes}
        \footnotesize
        \item VMOF W/O DS is the method without sampling directions based on Thompson sampling, VMOF W/O VO is the method without directions fine-tuning method.
        \item The Best Result in Each Row is Highlighted in Bold.
    \end{tablenotes}
  \end{threeparttable}
  \label{tab:ablation}%
\end{table}%

In order to verify the effectiveness of directions sampling and directions fine-tuning, we conduct an ablation experiment, which is reported in Table~\ref{tab:ablation}. We designed two variants of the VMOF algorithm, one without directions sampling component (W/O DS), and one without directions fine-tuning component (W/O DF). The experiments are carried out on the tri-objective LSMOP problems, and the dimensions of the decision variables vary from 10,000 to 500,000.
\par
First of all, it can be seen from Table~\ref{tab:ablation} that both variants of VMOF have the best results in some test instances, which shows that both components contribute to the improvement of the VMOF algorithm. 
Secondly, it can be seen from the final results that VMOF obtains the best results in 22 of 36 test cases, whilst the ablated variants obtain the best results in 16 of the 36 (two test cases are tied). Of these 16, 13 best results are obtained without direction sampling, and 5 are obtained without fine-tuning (again two test cases are tied). So, overall, it seems that whilst the picture is not completely clear, nevertheless, the overall VMOF, featuring both direction sampling and fine-tuning, is the best overall.
\subsection{Parameter Sensitivity Analysis}
\begin{figure}[htbp]
  \centering
  \includegraphics[width=1\hsize]{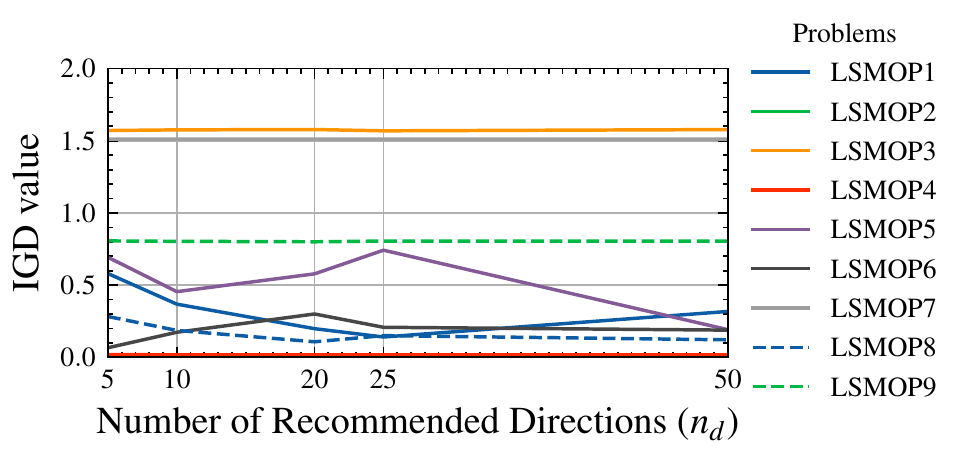}
    \caption{Statics of IGD results achieved by VMOF with different settings of the number of sampled directions $n_d$.}
    \label{fig:para}
\end{figure}

In the proposed VMOF, a parameter $n_d$ is used to directly control the number of directions recommended by Thompson sampling, and this parameter also indirectly controls the number of direction population during the evolution directions fine-tuning stage. To analyze the effect of this parameter on the performance of VMOF, experiments were conducted on a set of bi-objective LSMOP problems with 1,000 decision variables. The parameter $n_d$ is set to 5, 10, 20, 25, and 50, and the corresponding size of the diversity solution is 20, 10, 5, 4, and 2. The experimental results are shown in Fig.~\ref{fig:para}.  It can be observed that the performance of our proposed method is not significantly affected by different settings of $n_d$. This is especially obvious for the problems LSMOP2, LSMOP3, LSMOP4, LSMOP7, and LSMOP9, whilst for other problems, the performance has some fluctuation, but the result is still better compared to other state-of-the-art algorithms.
\par
This result can be attributed to the fact that direction sampling and direction fine-tuning are used simultaneously. When the number of recommended directions is insufficient, the size of direction population during the fine-tuning increases, thus compensating for the insufficient number of recommended directions. Conversely, when the size of direction population is insufficient, the number of recommended directions increases. Therefore, combined with the ablation experiments in the previous section, our proposed algorithm recommends direction and fine-tunes directions complementary to each other, improving the overall performance of the algorithm.
\subsection{Discussion}
We conducted a series of experiments to validate the effectiveness of our proposed algorithm. Initially, we selected a set of representative MOEAs for comparison on VLSMOPs, VLSMOPs*, and LSMOPs, where our algorithm demonstrated superior performance on the majority of test instances. Subsequently, we presented visualizations of IGD variations and time complexity to affirm the efficiency of our proposed framework. Lastly, we carried out ablation experiments and a parameter sensitivity analysis on direction sampling and fine-tuning. The outcomes of these experiments confirmed that both components enhance the search results and that our method exhibits robustness to these hyper-parameters.
\section{Conclusion and Future Work}
\label{sec:cl-fw}
MOPs with an increasing number of decision variables can be found widely in industrial, economic, and scientific research. In this work, we formalize the notion of very large-scale multiobjective optimization problems with decision variables of more than 100,000, highlighting that these are actually quite prevalent in the real world. Accordingly, we propose a framework based on direction sampling and fine-tuning to solve VLSMOPs. The algorithm adopts Thompson sampling to sample and recommend the direction that can guide the evolution of the population in the high-dimensional search space, and adopts the directions fine-tuning method to fine-tune the directions for each solution, so as to effectively solve very large-scale problems.
\par
To thoroughly comprehend and validate the effectiveness of the proposed VMOF, we conducted a comprehensive analysis of the framework and benchmarked the algorithm against state-of-the-art algorithms across a broad spectrum of instances. The experimental outcomes substantiate that the strategies of sampling and fine-tuning directions are particularly efficacious in tackling VLSMOPs. Future work could involve the incorporation of advanced techniques \cite{9097186, 9199822, ANG20101302, 7906564} for solving MOPs, which remains a worthwhile avenue of exploration. Furthermore, given the extensive practical applicability of such very large-scale problems, this significant issue warrants further investigation.

\section*{Acknowledgment}
This work was supported by the National Natural Science Foundation of China (No. 62276222), and in part by the public technology service platform project of Xiamen City (No.3502Z20231043).

\bibliographystyle{IEEEtran}
\bibliography{VMOF}

\begin{IEEEbiography}
[{\includegraphics[width=1in,height=1.25in,clip,keepaspectratio]{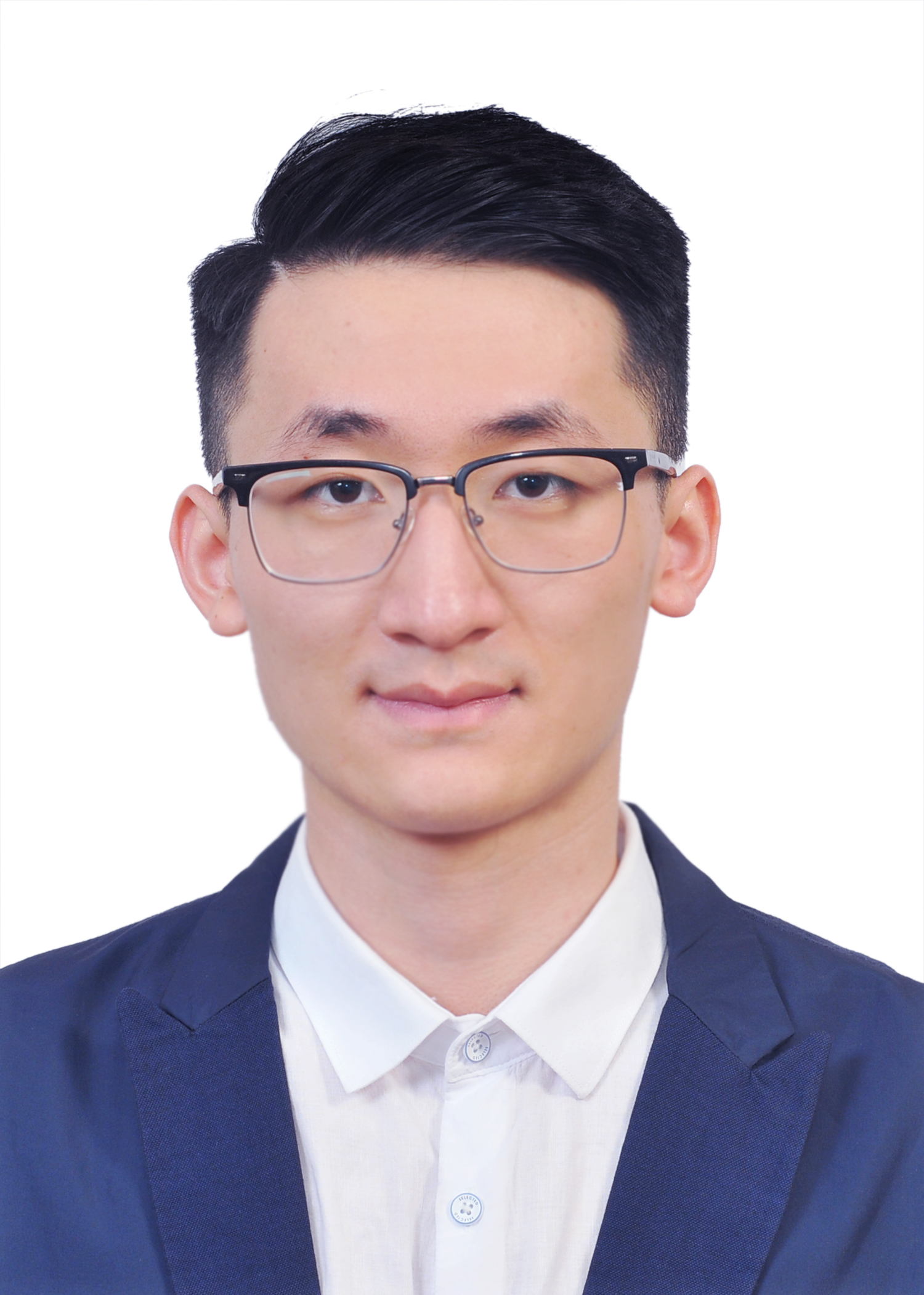}}]
{Haokai Hong} received both his Bachelor’s and Master’s degrees in Computer Science from Xiamen University, China, in 2020 and 2023, respectively. He is presently a Ph.D. candidate in the Department of Computing at The Hong Kong Polytechnic University, Hong Kong SAR.
\par
His research primarily focuses on the fields of evolutionary computation and machine learning.
\end{IEEEbiography}

\begin{IEEEbiography}[{\includegraphics[width=1in,height=1.25in,clip,keepaspectratio]{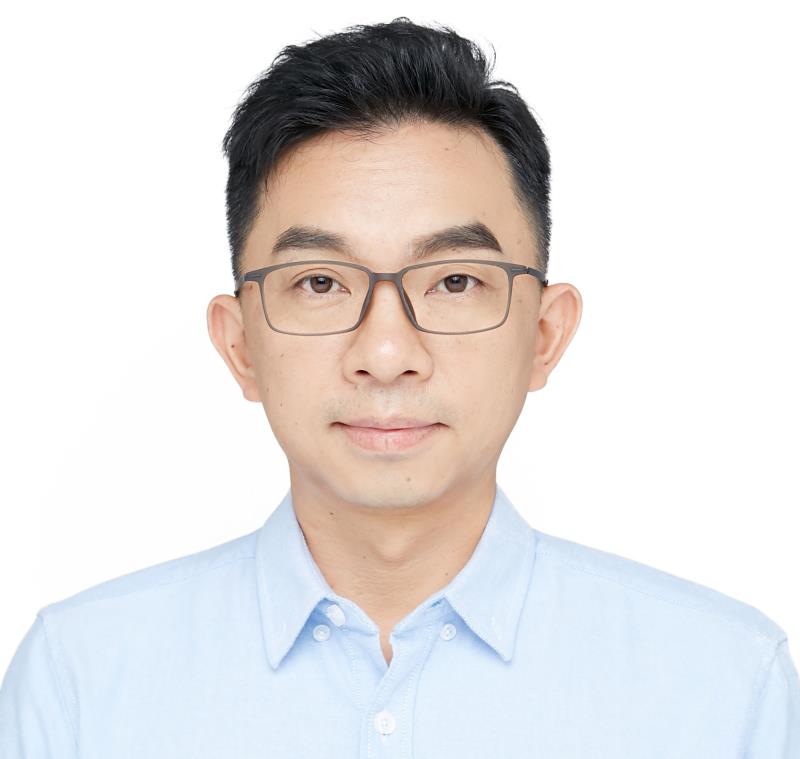}}]{Min Jiang}
`(M'11-SM'12) received his bachelor and Ph.D. degrees in computer science from Wuhan University, China, in 2001 and 2007, respectively. Subsequently as a postdoc in the Department of Mathematics of Xiamen University. Currently he is a professor in the Department of Artificial Intelligence, Xiamen University. His main research interests are Machine Learning, Computational Intelligence and the interplay between them. Dr. Jiang is currently serving as associate editors for the \textit{IEEE Transactions on Evolutionary Computation}, \textit{IEEE Transactions on Neural Networks and Learning Systems}, \textit{IEEE Computational Intelligence Magazine } and \textit{IEEE Transactions on Cognitive and Developmental Systems}. He is the Chair of IEEE CIS Xiamen Chapter.
\end{IEEEbiography}

\begin{IEEEbiography}[{\includegraphics[width=1in,height=1.25in,clip,keepaspectratio]{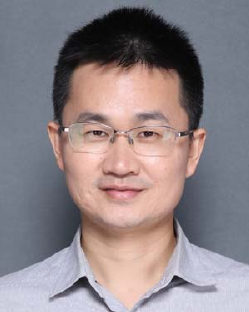}}]{Qiuzhen Lin} (Member, IEEE) received the B.S. degree from Zhaoqing University, Zhaoqing, China, in 2007, the M.S. degree from Shenzhen University, Shenzhen, China, in 2010, and the Ph.D. degree from the Department of Electronic Engineering, City University of Hong Kong, Hong Kong, in 2014. He is currently an Associate Professor with the College of Computer Science and Software Engineering, Shenzhen University. He has authored or coauthored more than 100 research papers since 2008. His research interests include artificial immune systems, multiobjective optimization, and dynamic systems. He is an Associate Editor for  \textit{IEEE Transactions on Evolutionary Computation} and  \textit{IEEE Transactions on Emerging Topics in Computational Intelligence}.
\end{IEEEbiography}

\begin{IEEEbiography}[{\includegraphics[width=1in,height=1.25in,clip,keepaspectratio]{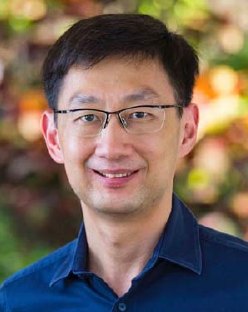} }]{Kay Chen Tan} (Fellow, IEEE) received the B.Eng. degree (first-class Hons.) and the Ph.D. degree from the University of Glasgow, Glasgow, U.K., in 1994 and 1997, respectively. He is currently a Chair Professor (Computational Intelligence) of the Department of Computing, the Hong Kong Polytechnic University, Hong Kong. He has authored or coauthored more than 300 refereed articles and seven books. Prof. Tan is currently the Vice-President (Publications) of IEEE Computational Intelligence Society, USA. He was the Editor-in-Chief of the \textit{IEEE Computational
Intelligence Magazine} from 2010 to 2013 and \textit{IEEE Transaction On Evolutionary Computation} from 2015 to 2020, and is currently the Editorial Board Member for more than ten journals. He is the Chief Co-Editor of Springer Book Series on \textit{Machine Learning: Foundations, Methodologies, and Applications}.
\end{IEEEbiography}

\end{document}